\documentclass[a4paper,journal,twocolumn]{IEEEtran}
\usepackage{amsmath,amsfonts}
\usepackage{algorithmic}
\usepackage{algorithm}
\usepackage{array}
\usepackage{color}
\usepackage{subcaption}
\usepackage{bm}
\usepackage{amssymb}
\usepackage{textcomp}
\usepackage{float}
\usepackage{url}
\usepackage{verbatim}
\usepackage{graphicx}
\usepackage{cite}
\usepackage{geometry}
\usepackage{graphicx}
\usepackage{tikz}
\usepackage{multirow} 
\usepackage{makecell} 
\newcolumntype{L}[1]{>{\raggedright\arraybackslash}p{#1}}
\newcolumntype{C}[1]{>{\centering\arraybackslash}p{#1}}
\usepackage[table,svgnames]{xcolor}
\usepackage[algo2e,ruled,vlined]{algorithm2e}
\usepackage{booktabs}
\usepackage{tabularx}
\usepackage[table]{xcolor}
\usepackage[section]{placeins}

\usepackage{booktabs}
\usepackage{graphicx}
\usepackage{makecell}
\usepackage{array}

\usepackage[colorlinks=true, linkcolor=blue, citecolor=green, urlcolor=red]{hyperref}
\newcommand{\RNum}[1]{\uppercase\expandafter{\romannumeral #1\relax}}

\definecolor{tblgray}{gray}{0.9}
\begin{document}
\title{Lightweight Adaptive Feature Composition for Heterogeneous Downstream Adaptation of Wireless Foundation Models}
	
	\author{Yuxuan Shi, Tingting Yang \IEEEmembership{Senior Member,~IEEE}, Li Sun \IEEEmembership{Senior Member,~IEEE}, Kangning Ma, Liwen Jing \IEEEmembership{Member,~IEEE}, Yuwei Wang, and Mengfan Zheng,
		\vspace{-0.3cm}
		\thanks{(corresponding author: Tingting Yang)
			
			Yuxuan Shi, Tingting Yang, Li Sun, Liwen Jing, Yuwei Wang, and Mengfan Zheng are with the Department of Broadband Communication, Pengcheng Laboratory, Shenzhen,
			518000, China (e-mail: shiyx01@pcl.ac.cn, yangtt@pcl.ac.cn, sunl03@pcl.ac.cn, jinglw@pcl.ac.cn, wangyw03@pcl.ac.cn, and zh.mengfan@gmail.com).
			
			Kangning Ma is with Purple Mountain Laboratories, Nanjing 211111, China (e-mail: makangning@pmlabs.com.cn).
			
			The relative repository is available at https://github.com/YXsjtu/RAFC-Routing-Adapter-for-Feature-Composition.
		}
	}
	
	
	\maketitle
	\pagestyle{empty}
	\thispagestyle{empty}
\begin{abstract}
	Mobile systems increasingly rely on heterogeneous learning-enabled wireless functions, for which separate task-specific models incur redundant training and model-management overhead. Wireless foundation models (WFMs) enable these functions to share a pretrained backbone, but existing adaptation either updates the backbone per task or relies on an inflexible final-layer representation. We observe that
	intermediate WFM layers exhibit distinct depth-dependent correlation
	structures. Based on this observation, we propose a Routing Adapter for Feature Composition (RAFC), which summarizes selected hidden states into compact descriptors, generates task- and sample-dependent routing weights, and combines the original full-resolution features without updating the pretrained backbone. Its gains arise from giving each task adaptive access to complementary representations across multiple depths rather than restricting it to the final layer. Experiments across four task categories and three WFM backbones show improvements in every evaluated backbone--task pair, with mean relative gains across the four task categories ranging from 2.9\% to 35.9\%, while RAFC adds only
	0.003M--0.049M trainable parameters across the evaluated backbones. Experiments on a portable software-defined radio testbed further demonstrate improved end-to-end receiver robustness, particularly under partial-band co-channel
	interference. These results establish RAFC as a lightweight, model-agnostic,
	and interpretable adaptation interface for WFMs in mobile systems.
\end{abstract}
	
	
	\section{Introduction}
		\IEEEPARstart{A}{s} mobile systems evolve toward increasingly intelligent and
		integrated operation, learning-based models are being adopted across channel
		processing, radio-resource management, and wireless sensing
		~\cite{Arvinte2023ScoreCE,Luan2023Channelformer,Vuckovic2024PARAMOUNT,
			Zhou2022MIMOOFDM,Park2026ResourceBeam,Zhang2026CSILocalization,jing2026signal}. Most existing methods follow a
		one-model-per-task paradigm, which can provide strong task-specific performance
		but repeatedly learns related wireless structures and requires separate model
		training, storage, and maintenance. These costs increase as the number of
		wireless functions and operating conditions grows. Wireless foundation models
		(WFMs) have therefore emerged as a scalable alternative
		~\cite{shen2024large,liang2026large,chen2024big}. By pretraining on extensive
		wireless data, a WFM provides a shared representation backbone that can be
		reused across heterogeneous downstream functions, reducing the need to maintain
		a separate backbone for every task.

		Owing to the remarkable success of the Transformer architecture in natural language processing and computer vision, recent WFMs have largely adopted Transformer-based backbones to model high-dimensional wireless observations. Existing studies can be broadly categorized into two lines. The first focuses on fine-tuning existing generative models through post-training methods for channel modeling, prediction, and downstream adaptation, as exemplified by ChannelGPT and LLM4CP~\cite{ChannelGPT,LLM4CP,FoundationModelSoM, liu2025llm4wm, kim2026large}. These approaches leverage pretrained large language models to learn channel structures and time--frequency correlations. The second line develops native wireless representation-learning strategies, with representative works including LWM, WirelessGPT, and WiMAE~\cite{WirelessGPT,LWM,WiFo,WiFo2,MultiTaskWFM,ContraWiMAE,WiFoMiSAC,Zhao2026CSIBERT2}. These studies emphasize native pretraining frameworks, general-purpose wireless representations, and multi-task transfer for communication and sensing. Despite these advances, most studies focus on backbone architectures, pretraining objectives, or task-specific adaptation. How to use a pretrained WFM efficiently and systematically for multiple heterogeneous downstream tasks remains insufficiently explored.
	
		Consequently, the central challenge shifts from learning a more capable backbone to determining how heterogeneous tasks should access and reuse the representations already encoded by a frozen WFM. Existing adaptation strategies present a practical trade-off. Full or
		parameter-efficient fine-tuning can modify the backbone for a specific target,
		but requires task-dependent updates and additional model management
		~\cite{houlsby2019parameter,hu2022lora}. For a mobile system that supports
		multiple wireless functions, maintaining a separate backbone variant for each
		task increases model storage, update, and switching overhead. Frozen-backbone
		adaptation is therefore attractive because the pretrained model can be shared
		without modification. Nevertheless, it commonly connects the task head only to
		the final encoder layer. This fixed interface assumes that one representation
		depth is suitable for every objective, although wireless tasks require different
		mixtures of physical information. For example, channel reconstruction favors local amplitude and phase details, whereas some decision-oriented tasks rely more on global channel-level quantities or states. Moreover, relying exclusively on the final layer can be restrictive because Transformer representations tend to become increasingly homogenized with depth,
		potentially weakening the token-level distinctions required by some downstream
		tasks~\cite{Nguyen2023OverSmoothing,Wang2022AntiOversmoothing,
			Jawahar2019BERTStructure,Rogers2020BERTology}. The resulting
		bottleneck is therefore a mismatch between a fixed final-layer representation
		and heterogeneous task requirements, rather than simply insufficient backbone
		capacity.
	
	In fact, layer aggregation has been explored in language and vision models. For instance, ELMo~\cite{Peters2018ELMo} employs task-specific scalar weighting over different layers, demonstrating that layer-wise features encapsulate complementary knowledge. Subsequent studies have further explored hierarchical aggregation, encoder fusion, and multi-scale feature alignment (e.g., AdapterFusion, ViT-Adapter, ViT-CoMer) to enhance task adaptation and modality disentanglement~\cite{Dou2018DeepRepresentations, Liu2021EncoderFusion, ViTAdapter, ViTCoMer, ElNokrashy2024DWAtt, AdapterFusion, TaskCustomizedMoA}. These studies consistently suggest that hidden representations at varying depths are not redundant but highly complementary. However, directly applying these mechanisms to WFMs encounters two major problems. First, existing fusion architectures (e.g., heavy cross-attention \cite{ViTAdapter, ViTCoMer} or dense multi-scale pooling \cite{Dou2018DeepRepresentations, Liu2021EncoderFusion}) typically introduce substantial trainable parameters, violating the stringent lightweight and low-latency requirements of wireless systems. Second, most of these multi-layer fusion techniques are intrinsically designed and optimized for individual, predefined downstream objectives, thereby tightly coupling the aggregated features to a single task. This fragmented approach does not provide a common and reusable representation interface for adapting a shared WFM backbone to heterogeneous downstream tasks. Consequently, a practical WFM interface should provide a lightweight, reusable, and interpretable feature-composition mechanism for heterogeneous downstream adaptation.

	To overcome these architectural bottlenecks, we propose a unified adaptive feature-composition framework that shifts downstream adaptation from single-layer extraction to dynamic multi-layer aggregation. Here, ``unified'' refers to a common feature-composition interface used across task types. The framework selects representations spanning shallow, intermediate, and deep layers, which retain fine-grained channel details and progressively broader channel context. These latent maps are fed into a lightweight \textit{Routing Adapter for Feature Composition (RAFC)}, which estimates the contribution of each selected WFM layer and aggregates the features into a task-specific representation. Downstream models can therefore exploit information preserved at multiple depths instead of relying on a fixed final-layer output. The framework offers three practical advantages: it is model agnostic and compatible with different WFMs and downstream architectures; it adds little computational overhead through simple pooling and compact neural networks; and its learned layer weights provide an interpretable view of feature reuse across wireless tasks. The main contributions are summarized as follows:

	\begin{itemize}
		\item We identify the fixed-depth representation mismatch in frozen WFM
		adaptation and empirically characterize depth-dependent token-correlation
		structures across different WFM encoders. The results show that representations
		at different depths exhibit distinct correlation ranges along the tokenized
		temporal, frequency, and spatial dimensions.

		\item We develop RAFC, a lightweight and model-agnostic module that
		separates low-dimensional routing decisions from full-resolution feature
		composition. Its compact descriptors and bottleneck network provide
		task- and sample-dependent adaptation with linear complexity in the number of
		selected layers and tokens, allowing multiple wireless functions to share a
		frozen WFM without maintaining task-specific backbone copies. The learned routing weights further reveal task-
		and sample-dependent layer preferences in WFMs.
				
		\item We establish a portable SDR-based indoor over-the-air validation
		platform whose wireless endpoints can be relocated across office and corridor
		locations. It supports end-to-end evaluation of WFM-based and conventional
		methods under location-dependent propagation, hardware impairments, and
		controlled interference.
	\end{itemize}

	Across channel estimation, channel prediction, beam prediction, and urban
	localization, RAFC improves every evaluated backbone--task pair. For WGPT, it
	improves the mean over seven signal-to-noise ratio (SNR) levels by 1.61~dB and 1.44~dB for channel estimation and
	channel prediction, respectively, raises 64-beam Top-1 accuracy by 1.88
	percentage points, and reduces mean localization error from 6.38 to 4.09~m.
	These changes correspond to relative improvements in mean performance between 2.9\% and
	35.9\%, while RAFC adds 0.003M--0.049M trainable parameters across the
	evaluated backbones. Comparisons with
	parameter-matched and full fine-tuning, together with layer-selection and
	pooling ablations, establish its performance--complexity trade-off. Finally,
	experiments on the portable SDR platform show that RAFC preserves nominal link
	performance and provides the most stable block error rate (BLER) under partial-band co-channel
	interference, demonstrating that the representation gains extend beyond
	offline task metrics.

\section{System Model and Problem Formulation}
\label{sec:problem}
	\begin{figure*}[t]
	\centering
	\includegraphics[width=0.8\linewidth]{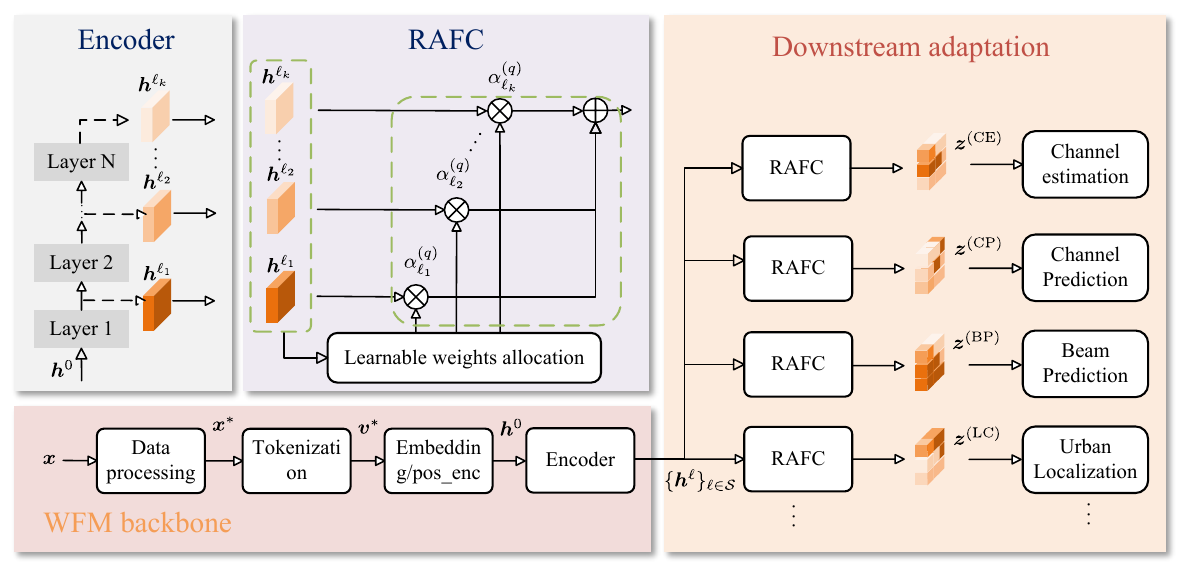}
	\captionsetup{font={small}}
	\caption{Overall depiction of the proposed unified adaptive feature-composition framework. The frozen WFM backbone provides multi-level CSI representations, while RAFC dynamically aggregates task-relevant layer-wise features to generate customized representations for different downstream tasks.}
	\label{framework}
\end{figure*}

This section presents the channel state information (CSI) observation model for a multiple-input multiple-output orthogonal frequency-division multiplexing (MIMO-OFDM) system, abstracts the hierarchical
representations, and formulates downstream adaptation
under a frozen backbone.

\subsection{MIMO-OFDM Signal and CSI Observation Model}
\label{subsec:system_model}

Consider a wideband MIMO-OFDM system with $N_t$ transmit antennas, $N_r$
receive antennas, $N_{\mathrm{sc}}$ subcarriers, and $N_s$ consecutive OFDM
symbols. At subcarrier $k$ and OFDM symbol $m$, the received signal is
\begin{equation}
	\bm{r}[k,m]
	=
	\bm{H}[k,m]\bm{s}[k,m]
	+
	\bm{n}[k,m],
	\label{eq:received_signal}
\end{equation}
where $\bm{s}[k,m]\in\mathbb{C}^{N_t}$ is the transmitted signal,
$\bm{H}[k,m]\in\mathbb{C}^{N_r\times N_t}$ is the frequency-domain
channel matrix, and
$\bm{n}[k,m]\sim\mathcal{CN}(\bm{0},\sigma_n^2\bm{I}_{N_r})$ is additive
noise. Stacking the channel matrices over all subcarriers and OFDM symbols
yields the CSI tensor
$\bm{H}_t\in\mathbb{C}^{N_r\times N_t\times N_{\mathrm{sc}}\times N_s}$
for slot $t$.

At pilot resource elements, least-squares (LS) estimation is applied to Eq.~\eqref{eq:received_signal}.
The resulting sparse CSI observation is represented as
\begin{equation}
	\widetilde{\bm{H}}_t
	=
	\bm{M}_t\odot\left(\bm{H}_t+\bm{E}_t\right),
	\label{eq:sparse_csi}
\end{equation}
where $\bm{M}_t\in\{0,1\}^{N_{\mathrm{sc}}\times N_s}$ is the pilot mask,
broadcast over the antenna dimensions, $\bm{E}_t$ denotes the LS estimation
error at observed positions, and $\odot$ is element-wise multiplication.
This representation covers both sparse noisy CSI and full CSI by selecting
the corresponding mask and noise level.

\subsection{Preprocessing and Hierarchical Representations}
\label{subsec:hierarchical_representation}

For a downstream task $q$, let $\bm{x}^{(q)}$ denote its CSI observation.
A backbone-specific operator $\mathcal{P}(\cdot)$ performs normalization,
real-valued decomposition, and tokenization. The resulting tokens are
embedded as
\begin{equation}
	\bm{h}^{0}
	=
	\mathcal{G}
	\left(
		\mathcal{P}(\bm{x}^{(q)})
	\right)
	\in
	\mathbb{R}^{L\times D},
	\label{eq:wfm_input}
\end{equation}
where $L$ and $D$ are the token length and hidden dimension, respectively. A pretrained WFM with $N$ Transformer layers then produces
\begin{equation}
	\bm{h}^{\ell}
	=
	\mathcal{E}^{\ell}_{\bm{\theta}^{\ell}}
	\left(
	\bm{h}^{\ell-1}
	\right),
	\qquad
	\ell=1,\ldots,N,
	\label{eq:wfm_layers}
\end{equation}
where $\bm{\theta}^{\ell}$ contains the parameters of layer $\ell$. All
backbone parameters $\bm{\theta}=\{\bm{\theta}^{1},\ldots,\bm{\theta}^{N}\}$
remain frozen during downstream adaptation. For a selected layer set
$\mathcal{S}=\{\ell_1,\ldots,\ell_K\}\subseteq\{1,\ldots,N\}$, the WFM
exposes the hierarchical feature pool
\begin{equation}
	\mathcal{H}_{\mathcal{S}}(\bm{x}^{(q)})
	=
	\left\{\bm{h}^{\ell_j}\right\}_{j=1}^{K}.
	\label{eq:hierarchical_pool}
\end{equation}
Transformer depth progressively changes the receptive field and abstraction
of the tokens. Consequently, the elements of Eq.~\eqref{eq:hierarchical_pool}
provide complementary representations rather than interchangeable copies.

\subsection{Heterogeneous Downstream Adaptation}
\label{subsec:adaptation_problem}

We consider four representative downstream tasks: channel estimation (CE),
channel prediction (CP), beam prediction (BP), and localization (Loc.). Their
combination reflects the heterogeneous computing functions required by a mobile
link, from recovering the current channel and anticipating its evolution to
supporting radio-resource decisions and location-aware services. Their
input--target pairs are summarized as
\begin{equation}
	\begin{aligned}
		\mathrm{CE}:  && \bm{x}^{(\mathrm{CE})}&=\widetilde{\bm{H}}_t,
		&\bm{y}^{(\mathrm{CE})}&=\bm{H}_t,\\
		\mathrm{CP}:  && \bm{x}^{(\mathrm{CP})}&=\widetilde{\bm{H}}_t,
		&\bm{y}^{(\mathrm{CP})}&=\bm{H}_{t+1},\\
		\mathrm{BP}:  && \bm{x}^{(\mathrm{BP})}&=\widetilde{\bm{H}}_t,
		&\bm{y}^{(\mathrm{BP})}&=c_{t+1}^{\star},\\
		\mathrm{Loc.}:&& \bm{x}^{(\mathrm{Loc.})}&=\bm{H}_t,
		&\bm{y}^{(\mathrm{Loc.})}&=\bm{p}_t,
	\end{aligned}
	\label{eq:task_definitions}
\end{equation}
where $\bm{p}_t$ is the user equipment (UE) position. Given a transmit codebook
$\mathcal{C}=\{\bm{f}_c\}_{c=1}^{|\mathcal{C}|}$, the optimal beam index is
\begin{equation}
	c_{t+1}^{\star}
	=
	\arg\max_{c\in\{1,\ldots,|\mathcal{C}|\}}
	\sum_{k,m}\left\|\bm{H}_{t+1}[k,m]\bm{f}_c\right\|_2^2.
	\label{eq:optimal_beam}
\end{equation}
Thus, CE and CP recover signal-level channel states, whereas BP and
localization infer system-level quantities.

For task $q$, let
$\mathcal{D}^{(q)}=\{(\bm{x}_{i}^{(q)},\bm{y}_{i}^{(q)})\}_{i=1}^{I_q}$.
Conventional frozen-backbone adaptation predicts
$\widehat{\bm{y}}^{(q)}=\mathcal{Q}_{\bm{\omega}^{(q)}}(\bm{h}^{N})$ using
only the final-layer representation. This fixed interface restricts every
task to the same representation depth, despite their distinct requirements
for local channel details and global spatial structure.

We instead seek a lightweight, task-specific composition function that
operates on the feature pool in Eq.~\eqref{eq:hierarchical_pool}:
\begin{equation}
	\bm{z}^{(q)}
	=
	\mathcal{A}_{\bm{\phi}^{(q)}}
	\left(
		\mathcal{H}_{\mathcal{S}}(\bm{x}^{(q)})
	\right),
	\qquad
	\widehat{\bm{y}}^{(q)}
	=
	\mathcal{Q}_{\bm{\omega}^{(q)}}
	\left(
	\bm{z}^{(q)}
	\right),
	\label{eq:adaptation_interface}
\end{equation}
where $\bm{\phi}^{(q)}$ and $\bm{\omega}^{(q)}$ parameterize the composition
function and task head, respectively. They are optimized as
\begin{equation}
	\min_{\bm{\phi}^{(q)},\,\bm{\omega}^{(q)}}
	\mathbb{E}_{(\bm{x},\bm{y})\sim\mathcal{D}^{(q)}}
	\left[
	\mathcal{L}^{(q)}
	\left(
	\widehat{\bm{y}}^{(q)},\bm{y}
	\right)
	\right],
	\qquad
	\bm{\theta}\ \text{frozen}.
	\label{eq:adaptation_objective}
\end{equation}
Here, $\mathcal{L}^{(q)}$ is a reconstruction, classification, or coordinate-regression
loss according to the task. The objective is therefore to improve task adaptation
without updating or replicating the pretrained backbone. The next section realizes
$\mathcal{A}_{\bm{\phi}^{(q)}}(\cdot)$ using the proposed RAFC.

\section{Routing Adapter for Feature Composition}
\label{sec:method}

\begin{figure*}[t]
	\centering
	\includegraphics[width=1\textwidth]{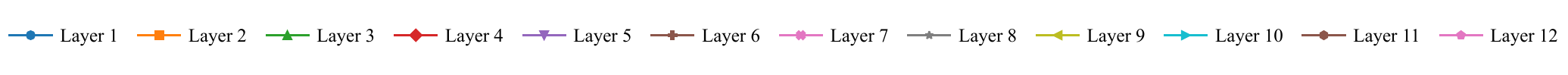}
	\\
	\begin{subfigure}{1\textwidth}
		\centering
		\includegraphics[width=\linewidth]{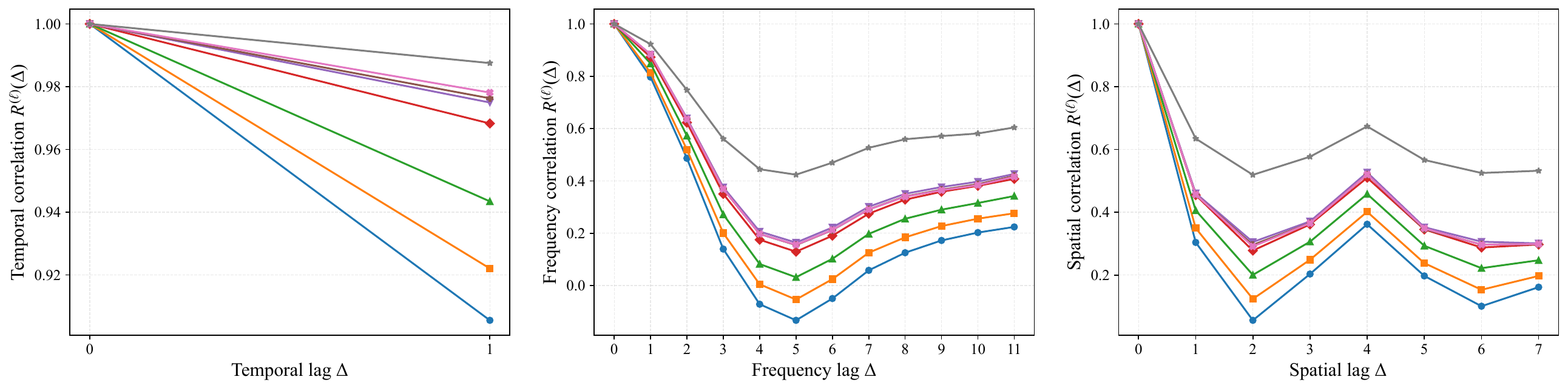}
		\caption{}
	\end{subfigure}
	\begin{subfigure}{0.66\textwidth}
		\centering
		\includegraphics[width=\linewidth]{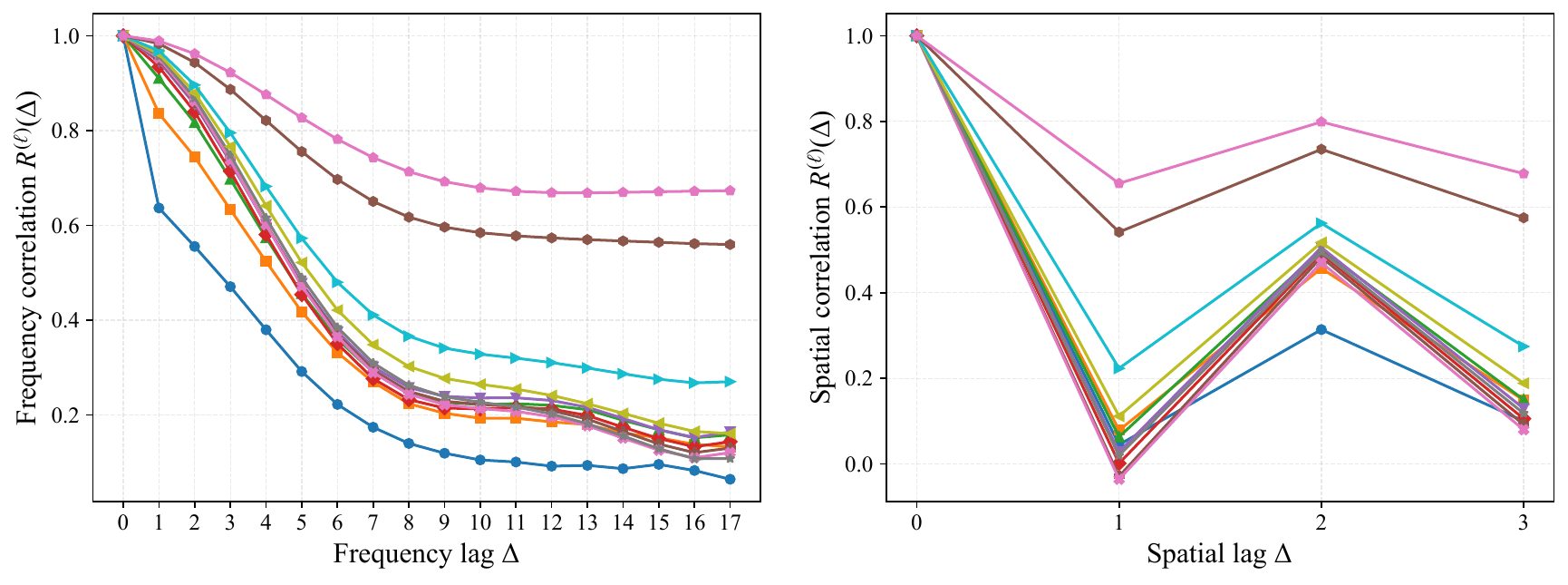}
		\caption{}
	\end{subfigure}
	
	\captionsetup{font={small}, justification=raggedright}
	\caption{Layer-wise token correlation under different physical spans $\Delta$:
			(a) temporal, frequency, and spatial spans for the eight-layer WGPT
			(the $(2,12,8)$ token grid contains 2 temporal, 12 frequency, and 8 spatial tokens);
			and (b) frequency and spatial spans for the twelve-layer LWM~v1.1
			(the $(18,4)$ token grid contains 18 frequency and 4 spatial tokens).}
	\label{layer}
\end{figure*}

The construction of RAFC follows a simple premise: multi-layer composition is
useful only if representations at different depths are not statistically redundant.
We therefore first characterize the depth-dependent correlation structures learned by
representative WFM encoders. Based on the observed correlation diversity, we then
introduce a lightweight composition mechanism that predicts task- and
sample-dependent layer weights while applying them to the original
full-resolution features. Finally, we describe how this mechanism supports
heterogeneous tasks with a shared frozen backbone and quantify its additional
complexity.

\subsection{Layer-Wise Feature Diversity in WFM Encoders}
\label{subsec:layer_diversity}

Although layer-wise hierarchies have been widely observed in language and vision
Transformers, their meaning in a WFM must be related to the physical organization
of CSI. We examine this question using WirelessGPT (WGPT)~\cite{WirelessGPT} and Large Wireless Model (LWM)~v1.1~\cite{LWM},
which adopt different tokenizations and encoder depths. WGPT organizes CSI tokens
over the temporal, frequency, and spatial dimensions, whereas LWM~v1.1 operates
on spatial--frequency channel matrices. Their comparison allows us to determine
whether depth-dependent behavior is specific to one architecture or persists
across WFM designs.

Let $\bm{h}^{\ell}_{i}$ denote the $i$-th token representation at layer $\ell$.
For a physical axis and token span $\Delta$, we measure the mean cosine
correlation
\begin{equation}
	R^{(\ell)}(\Delta)
	=
	\frac{1}{|\mathcal{P}_{\Delta}|}
	\sum_{(i,j)\in\mathcal{P}_{\Delta}}
	\frac{(\bm{h}^{\ell}_{i})^{\mathsf{T}}\bm{h}^{\ell}_{j}}
	{\|\bm{h}^{\ell}_{i}\|_{2}\|\bm{h}^{\ell}_{j}\|_{2}},
	\label{eq:physical_correlation}
\end{equation}
where $\mathcal{P}_{\Delta}$ contains token pairs separated by exactly $\Delta$
along that axis while the remaining coordinates are aligned. A rapid decrease in
$R^{(\ell)}(\Delta)$ means that token similarity diminishes quickly with physical
separation, whereas a slower decrease means that correlation persists over a
broader span.

Fig.~\ref{layer}(a) shows that the shallow WGPT layers exhibit the fastest
correlation decay along the temporal, frequency, and spatial dimensions. As depth
increases, token correlations persist over progressively larger spans.
Fig.~\ref{layer}(b) shows a similar depth-dependent pattern for the
spatial--frequency tokens of LWM~v1.1 despite its different tokenizer and encoder
depth. These results indicate that the observed correlation variation is not
limited to a single WFM architecture.

Importantly, these observations do not imply that either shallow or deep features
are universally superior, nor do the correlation profiles alone identify the
physical information encoded at each depth. Instead, they show that representations
from different encoder depths exhibit distinct statistical structures. Their
downstream relevance is evaluated empirically through the routing and ablation
results that follow. This depth-dependent representation diversity motivates
adaptive multi-layer composition.

\subsection{Lightweight Sample-Adaptive Feature Composition}
\label{subsec:RAFC}

\begin{figure}[t]
	\centering
	\includegraphics[width=0.9\linewidth]{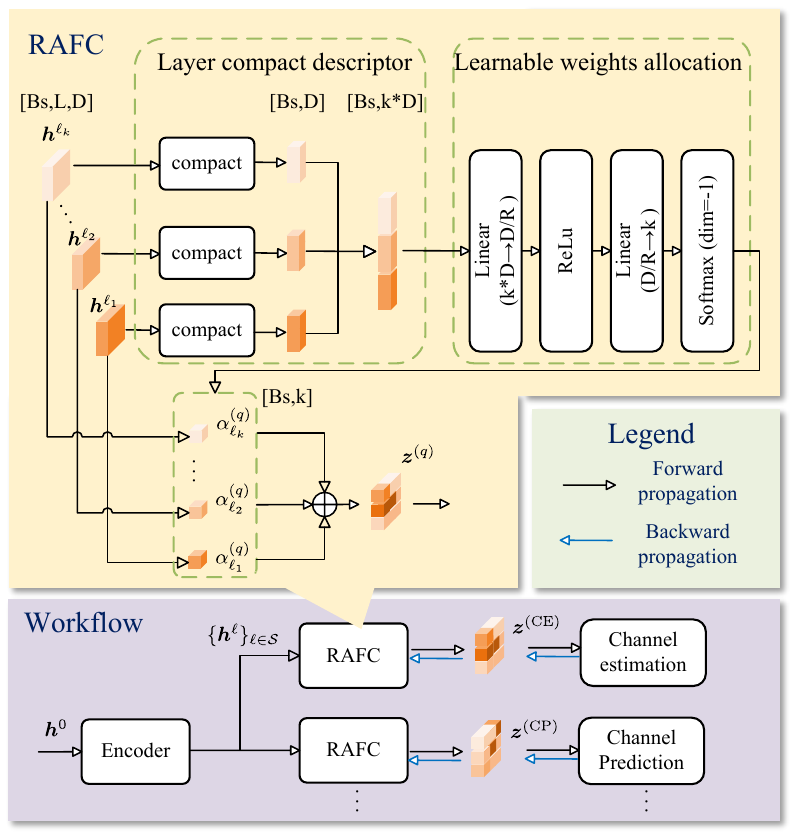}
	\captionsetup{font={small}}
	\caption{Architecture of RAFC.}
	\label{RAFC}
\end{figure}

Direct concatenation of multiple full token sequences expands the task-head input
from $D$ to $KD$, while dense cross-layer attention introduces additional
token-to-token interactions. Conversely, static averaging is inexpensive but
assigns the same layer preference to every task and channel realization. RAFC
avoids these two extremes by separating the low-dimensional routing decision from
the high-dimensional features being routed. As illustrated in Fig.~\ref{RAFC}, it
performs layer summarization, dynamic weight generation, and full-resolution
feature composition.

\emph{Layer summarization:}
For each selected layer $\ell_j\in\mathcal{S}$, RAFC first summarizes its token
sequence by \textbf{mean pooling}:
\begin{equation}
	\bm{u}_{\ell_j}
	=
	\frac{1}{L}\sum_{i=1}^{L}\bm{h}^{\ell_j}_{i}
	\in\mathbb{R}^{D},
	\qquad j=1,\ldots,K.
	\label{eq:layer_descriptor}
\end{equation}
This operation reduces each layer to a global descriptor while avoiding an
additional attention module. The descriptors are concatenated as
$\bm{u}=[\bm{u}_{\ell_1};\cdots;\bm{u}_{\ell_K}]\in\mathbb{R}^{KD}$.

\emph{Dynamic weight generation:}
A task-specific bottleneck multilayer perceptron (MLP) maps $\bm{u}$ to the routing distribution
\begin{equation}
	\bm{\alpha}^{(q)}(\bm{x})
	=
	\operatorname{softmax}\!\left(
	\bm{W}^{(q)}_2
	\sigma\!\left(\bm{W}^{(q)}_1\bm{u}+\bm{b}^{(q)}_1\right)
	+\bm{b}^{(q)}_2
	\right),
	\label{eq:routing_weights}
\end{equation}
where $\sigma(\cdot)$ is the rectified linear unit (ReLU),
$\bm{W}^{(q)}_1\in\mathbb{R}^{d_r\times KD}$,
$\bm{W}^{(q)}_2\in\mathbb{R}^{K\times d_r}$, and $d_r=D/R$ for reduction
ratio $R$. The output
$\bm{\alpha}^{(q)}=[\alpha^{(q)}_1,\ldots,\alpha^{(q)}_K]^{\mathsf T}$
satisfies $\alpha^{(q)}_j\geq0$ and $\sum_j\alpha^{(q)}_j=1$. Because the
weights depend on both the task-specific router and the current input, they encode
two forms of adaptation: the parameters of each router capture the persistent
depth preference of task $q$, while $\bm{u}$ allows that preference to vary with
the current channel realization.

\emph{Full-resolution feature composition:}
The routing weights are applied to the original feature maps
\begin{equation}
	\bm{z}^{(q)}
	=
	\sum_{j=1}^{K}\alpha^{(q)}_j(\bm{x})\bm{h}^{\ell_j}
	\in\mathbb{R}^{L\times D}.
	\label{eq:feature_composition}
\end{equation}
A single scalar is assigned to each layer and shared across its tokens. Hence,
the pooled descriptors determine only \emph{which depth} should contribute; they
do not replace the features delivered to the task head. Equation~\eqref{eq:feature_composition}
preserves the original $L\times D$ token organization and allows RAFC to retain
fine-grained information without dense cross-layer interaction.

\subsection{Task-Specific Adaptation}

The term \emph{unified adaptation} refers to a common composition interface for
all downstream tasks, rather than a single router shared by incompatible
objectives. Each task uses an independent RAFC module and task head, while the
pretrained WFM is shared and remains frozen. This separation allows heterogeneous
tasks to learn different routing policies without modifying or replicating the
backbone. For task $q$, only
$\bm{\phi}^{(q)}=\{\bm{W}^{(q)}_1,\bm{b}^{(q)}_1,
\bm{W}^{(q)}_2,\bm{b}^{(q)}_2\}$ and $\bm{\omega}^{(q)}$ are updated under
Eq.~\eqref{eq:adaptation_objective}; the WFM parameters $\bm{\theta}$ remain fixed.
Consequently, signal-level and system-level tasks can access the same hierarchical
feature pool through a shared procedure while learning task-appropriate mixtures.

During inference, one frozen-backbone pass produces the selected hidden states.
RAFC then evaluates Eqs.~\eqref{eq:layer_descriptor}--\eqref{eq:routing_weights},
forms Eq.~\eqref{eq:feature_composition}, and invokes the task head. No additional
Transformer block, iterative routing procedure, or backbone update is required.
The deployed model therefore preserves the stability and reuse benefits of a
frozen WFM, with only the small router and task head switched between tasks.

\subsection{Layer Selection and Complexity}
\label{selection}

The correlation profiles in Fig.~\ref{layer} suggest that useful diversity is
distributed across representation depth, whereas adjacent layers often exhibit
similar behavior. We therefore select one shallow, one intermediate, and one deep
layer to cover the hierarchy without routing every hidden state. Specifically,
$\mathcal{S}=\{1,5,8\}$ is used for the eight-layer WGPT, and
$\mathcal{S}=\{1,6,12\}$ is used for the twelve-layer backbones. The ablation
study evaluates these spaced selections against contiguous-layer and all-layer
alternatives.

With $K$ selected layers and bottleneck width $d_r$, the router contains
\begin{equation}
	P_{\mathrm{RAFC}}
	=
	KDd_r+Kd_r+d_r+K,
	\label{eq:rafc_parameters}
\end{equation}
trainable parameters. Its weight-generation cost is
$\mathcal{O}(KDd_r)$, while mean pooling and feature composition require
$\mathcal{O}(KLD)$ operations. Both costs scale linearly with token length and
the number of selected layers. Moreover, the weighted sum returns an $L\times D$
representation, so the downstream head does not expand with $K$. RAFC therefore
exposes complementary WFM depths without quadratic cross-token interaction or
incurring the parameter and storage costs of backbone fine-tuning. In summary, the proposed
design exploits the depth-dependent statistical diversity observed in Fig.~\ref{layer} into a
lightweight task- and sample-adaptive interface for heterogeneous downstream
adaptation.

	\section{Experiments}
	
	\begin{table*}[t]
		\centering
		\caption{Pretraining and downstream experimental configurations.}
		\label{tab:experimental_configurations}
		\begin{subtable}{\textwidth}
			\centering
			\caption{Pretraining dataset, model, and optimization configurations.}
			\label{tab:pretraining_details}
			\renewcommand{\arraystretch}{1.15}
			\setlength{\tabcolsep}{4.0pt}
			\scriptsize
			
			\begin{tabular}{
					@{}
					L{0.145\textwidth}
					C{0.145\textwidth}
					L{0.180\textwidth}
					C{0.130\textwidth}
					C{0.130\textwidth}
					C{0.145\textwidth}
					@{}
				}
				\toprule
				
				\rowcolor{gray!20}
				\multicolumn{2}{c}{\textbf{Dataset Configurations}}
				&
				\multicolumn{4}{c}{\textbf{Models and Optimization}}
				\\
				
				\cmidrule(lr){1-2}
				\cmidrule(lr){3-6}
				
				\textbf{Item}
				& \textbf{\makecell{DeepMIMO\\(16 Scenarios)}}
				& \textbf{Item}
				& \textbf{WGPT}
				& \textbf{LWMv1.1}
				& \textbf{ContraWiMAE}
				\\
				
				\midrule
				
				Center frequency
				& 3.5 GHz
				& Pretraining objective
				& Masked autoencoding
				& Masked autoencoding
				& \makecell{Masked autoencoding +\\InfoNCE}
				\\
				
				\# base stations
				& 3
				& \# encoder/decoder layers
				& 8/linear head
				& 12/linear head
				& 12/4
				\\
				
				\# subcarriers
				& \makecell{144 (12 physical\\resource blocks)}
				& \# encoder/decoder heads
				& 8/--
				& 8/--
				& 16/8
				\\
				
				Subcarrier spacing
				& 30 kHz
				& \# Hidden dimension
				& 256
				& 128
				& 64
				\\
				
				\# transmit antennas
				& \makecell{$4$: $\{4{\times}1,\,2{\times}2\}$}
				& Patch size over $\mathbb{C}$
				& $(2,6,7)$
				& $(4,4)$
				& $(16,1)$
				\\
				
				\# receive antennas
				& \makecell{$4$: $\{4{\times}1,\,2{\times}2\}$}
				& Masking structure
				& \makecell{
					Random: 50\%
				}
				& Random: 40\%
				& Random: 90\%
				\\
				
				\# OFDM symbols
				& 14
				& Batch size
				& 256
				& 128
				& 64
				\\
				
				Mobility (m/s)
				& \makecell{$[0,2]$; $[2,15]$;$[15,35]$}
				& Candidate blocks
				& $\{1,5,8\}$
				& $\{1,6,12\}$
				& $\{1,6,12\}$
				\\
				
				\# propagation paths
				& $\{10,15,25\}$
				& Input sequence size
				& \makecell{$(N,16,144,14)$}
				& \makecell{$(14N,16,144)$}
				& \makecell{$(14N,16,144)$}
				\\
				
				Train/validation split
				& 312k/78k
				& Trainable parameters
				& 6.80 M
				& 2.47 M
				& 0.57 M
				\\
				
				--
				& --
				& \textit{RAFC parameters}
				& \textit{0.049 M}
				& \textit{0.012 M}
				& \textit{0.003 M}
				\\
				
				\bottomrule
			\end{tabular}
			\begin{minipage}{\textwidth}
				\scriptsize
				\vspace{1mm}
				\textit{Note: InfoNCE denotes the information noise-contrastive estimation objective. During data generation, the transmit and receive array geometries are independently and uniformly selected from a $4\times1$ uniform linear array (ULA) and a $2\times2$ uniform planar array (UPA). The user velocity is uniformly selected from three intervals, i.e., $[0,2]$, $[2,15]$, and $[15,35]$~m/s, while the number of propagation paths is uniformly selected from $\{10,15,25\}$. WGPT jointly models all 14 OFDM symbols of each sample as a complete spatial--frequency--temporal CSI sequence, resulting in an input size of $(N,16,144,14)$. In contrast, LWMv1.1 and ContraWiMAE operate on two-dimensional channel matrices without explicitly modeling the temporal dimension, so the 14 OFDM symbols are unfolded into the sample dimension. Furthermore, the number of trainable RAFC parameters depends strongly on the output feature dimension of the upstream foundation model.} 
			\end{minipage}
		\end{subtable}
		
		\vspace{1.5mm}
		\begin{subtable}{\textwidth}
			\centering
			\caption{Downstream datasets, observation models, and task-head configurations.}
			\label{tab:downstream_config}
			\renewcommand{\arraystretch}{1.12}
			\setlength{\tabcolsep}{3.7pt}
			\scriptsize
			\begin{tabular}{@{}L{0.198\textwidth}C{0.180\textwidth}C{0.180\textwidth}C{0.180\textwidth}C{0.180\textwidth}@{}}
				\toprule
				\textbf{Item}
				& \textbf{Channel Estimation}
				& \textbf{Channel Prediction}
				& \textbf{Beam Prediction}
				& \textbf{Localization} \\
				\midrule
				\rowcolor{gray!20}
				\multicolumn{5}{c}{\textbf{Downstream datasets and observations}} \\
				\midrule
				DeepMIMO scenario(s)
				& City 9 (unseen)
				& City 9 (unseen)
				& City 9 (unseen)
				& City 16 (unseen) \\
				Carrier / subcarrier spacing
				& 3.5 GHz / 30 kHz
				& 3.5 GHz / 30 kHz
				& 3.5 GHz / 30 kHz
				& 3.5 GHz / 30 kHz \\
				MIMO / subcarriers
				& $4\times4$ / 72
				& $4\times4$ / 72
				& $4\times4$ / 72
				& $32\times2$ / 144 \\
				OFDM symbols
				& 14
				& 14 
				& 14 
				& 14 \\
				Input 
				& \makecell[c]{Sparse CSI}
				& \makecell[c]{Sparse CSI}
				& \makecell[c]{Sparse CSI}
				& Full CSI \\
				Target
				& Full CSI $\bm{H}_t$
				& \makecell[c]{Full CSI $\bm{H}_{t+1}$\\ in the next slot }
				& \makecell[c]{Optimal codeword\\ in the next slot}
				& UE position \\
				Split (train/val/test)
				& 8.4k / 1.8k / 1.8k
				& 8.4k / 1.8k / 1.8k
				& 8.4k / 1.8k / 1.8k
				& 31.5k / 6.75k / 6.75k \\
				Test SNR (dB)
				& \{0,5,10,15,20,25,30\}
				& \{0,5,10,15,20,25,30\}
				& \{0,5,10,15,20,25,30\}
				& Clean CSI \\
				\midrule
				\rowcolor{gray!20}
				\multicolumn{5}{c}{\textbf{Downstream heads and training protocol}} \\
				\midrule
				Backbone / RAFC
				& Frozen / trainable
				& Frozen / trainable
				& Frozen / trainable
				& Frozen / trainable \\
				Task head
				& 2-layer Transformer
				& 2-layer Transformer
				& 2-layer Transformer classifier
				& 2-layer Transformer regressor \\
				Hidden / heads / feed-forward
				& 256 / 8 / 512
				& 256 / 8 / 512
				& 256 / 8 / 512
				& 256 / 8 / 1024 \\
				Train/eval batch size
				& 256 / 128
				& 256 / 128
				& 256 / 128
				& 256 / 256 \\
				Max. epochs / patience
				& 1000 / 30
				& 1000 / 30
				& 500 / 30
				& 500 / 30 \\
				Learning rate
				& $10^{-4}$
				& $10^{-4}$
				& $10^{-4}$
				& $10^{-4}$ \\
				Validation criterion
				& \makecell{Normalized mean square\\error (NMSE)}
				& NMSE
				& Cross-entropy
				& Mean Euclidean error \\
				Reported metrics
				& NMSE
				& NMSE
				& \makecell[c]{Top-1 accuracy; macro-F1;\\weighted-F1}
				& \makecell[c]{Cumulative distribution\\function (CDF)}; MEE \\
				\bottomrule
			\end{tabular}
		\end{subtable}
	\end{table*}
	
	This section evaluates whether RAFC provides a general adaptation interface for
	a frozen WFM rather than a task-specific improvement tied to one backbone. We
	therefore examine both \emph{backbone generality}, i.e., whether RAFC remains
	effective under different WFM architectures and tokenizations, and \emph{cross-task
	applicability}, i.e., whether the same composition principle remains effective for signal-level
	reconstruction, temporal prediction, discrete beam selection, and spatial
	regression. We further analyze the learned routing policies, computational cost,
	and over-the-air performance. Unless stated otherwise, all learned results are
	reported over five independent runs with different random seeds. In each run,
	RAFC and the task head are randomly initialized and trained from scratch, while the pretrained WFM backbone remains frozen.

	\subsection{Experimental Protocol}
	
	\subsubsection{Datasets, Sampling Distributions, Tasks, and Metrics}
	The pretraining corpus contains 390k samples generated from 16 DeepMIMO
	scenarios and three base stations, of which 312k and 78k samples are used for
	training and validation, respectively. To avoid concentrating the corpus on a
	single channel regime, the transmitter and receiver array geometries are sampled
	independently and uniformly from a $4\times1$ ULA and a $2\times2$ UPA. The
	velocity category is sampled uniformly from $[0,2]$, $[2,15]$, and $[15,35]$
	m/s, and the number of paths is sampled uniformly from $\{10,15,25\}$. The
	resulting mixture covers different mobility, multipath, and array distributions
	under a common 3.5-GHz, 30-kHz subcarrier-spacing configuration. Further parameters are listed
	in Table~\ref{tab:pretraining_details}.

	Downstream samples are generated independently from unseen environments and are
	not included in the pretraining mixture. Channel estimation, channel prediction,
	and beam prediction use 12k samples from City~9 with an 8.4k/1.8k/1.8k
	train/validation/test split. These tasks use 72 subcarriers and sparse noisy CSI,
	but differ in their targets: the current full CSI, the next-slot full CSI, and
	the next-slot optimal codeword, respectively. Localization uses 45k samples from
	City~16 with a 31.5k/6.75k/6.75k split, 144 subcarriers, and full CSI as input.
	Consequently, evaluation introduces simultaneous shifts in propagation scene,
	sample distribution, observed bandwidth, input sparsity, and target semantics.

	Channel estimation and prediction are evaluated using NMSE in decibels. Beam
	prediction is evaluated by Top-1 accuracy, macro-F1, and weighted-F1 under
	32- and 64-entry codebooks; macro-F1 gives each class equal weight, whereas
	weighted-F1 weights each class by its number of test samples. Localization is evaluated using mean Euclidean error (MEE)
	and its empirical CDF. For the three noise-dependent tasks, the test SNR ranges
	from 0 to 30~dB in 5~dB increments.

	\subsubsection{Pretraining and Downstream Adaptation Protocol}
	All backbones are pretrained from random initialization on the common corpus
	described above. Each architecture retains its native masked-reconstruction
	objective, tokenizer, and masking pattern; the masking ratios and optimization
	batch sizes are reported in Table~\ref{tab:pretraining_details}. For downstream adaptation, the selected pretrained checkpoint is loaded and the
	entire WFM, including its embedding and Transformer layers, is frozen. A forward
	pass exposes the prescribed shallow, intermediate, and deep hidden states; only
	RAFC and the task head receive gradient updates. The batch sizes, patience,
	optimizer, and evaluation criteria are detailed in
	Table~\ref{tab:downstream_config}.

		\begin{figure}[t]
		\centering
		\includegraphics[width=0.85\linewidth,trim=2 2 2 2,clip]{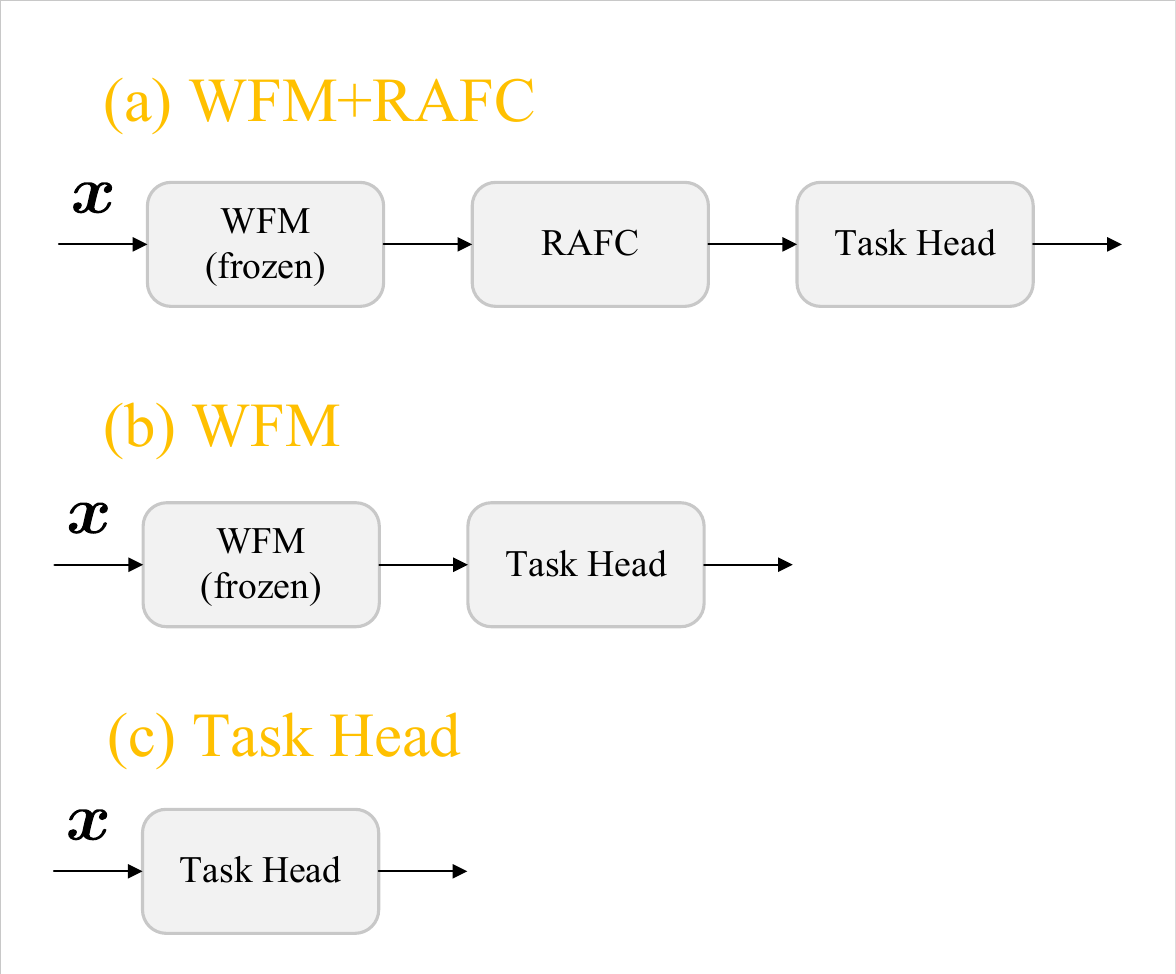}
		\captionsetup{font={small}, justification=raggedright}
		\caption{Clarification for the three baselines, where WFM $\in$ \{WGPT, LWM~v1.1, ContraWiMAE\}}
		\label{illust}
	\end{figure}
	
	\subsubsection{Baselines and Implementation Details}
	The baselines fall into two categories. The first comprises \textbf{WFM backbones:}\\
	\noindent \emph{WirelessGPT (WGPT)}~\cite{WirelessGPT} is a masked-autoencoding WFM that
	jointly tokenizes the spatial, frequency, and temporal dimensions of a CSI block.
	Its final encoder layer is connected to the task head to form the WGPT baseline;
	WGPT+RAFC instead exposes Layers $\{1,5,8\}$ while keeping the same frozen
	backbone and task head.

	\noindent\emph{LWM~v1.1}~\cite{LWM} is a lightweight masked wireless model whose encoder
	operates on two-dimensional channel matrices. It processes each OFDM symbol as
	an individual channel sample and uses the final layer for conventional frozen
	adaptation; LWM~v1.1+RAFC composes Layers $\{1,6,12\}$ before the same task head.

	\noindent\emph{ContraWiMAE}~\cite{ContraWiMAE} combines masked reconstruction with
	contrastive representation learning and uses a four-layer reconstruction decoder
	during pretraining. The decoder is discarded for downstream tasks, and the final
	encoder layer constitutes the baseline representation. ContraWiMAE+RAFC exposes
	Layers $\{1,6,12\}$ and learns only the router and task head.

	The second category comprises \textbf{task-specific conventional baselines:} \\
	\noindent The \emph{Task Head} baseline trains the same downstream architecture directly
	without pretrained WFM features. It isolates the contribution of representation
	pretraining while preserving the downstream capacity.

	\noindent \emph{LS Estimation} performs linear interpolation in the
	time--frequency domain based on the LS estimates at pilot positions~\cite{Coleri2002OFDM}, whereas \emph{DFT-LMMSE}, i.e., discrete Fourier transform-aided linear minimum mean square error estimation, exploits delay-domain
	sparsity and second-order channel statistics to interpolate the missing resource
	elements~\cite{Edfors1998OFDM}.
	These methods provide nonlearned channel-estimation references with different
	interpolation capabilities.

	\noindent \emph{Wiener-Sparse} applies Wiener filtering~\cite{DuelHallen2000Prediction} to the same sparse CSI available to the learned methods. \emph{Wiener-Perfect} is supplied
	with complete current CSI and therefore serves as an idealized reference rather
	than a practical competitor for channel prediction.

	\noindent \emph{MUSIC--WKNN} extracts angle-related features using multiple signal classification (MUSIC)~\cite{Schmidt1986MUSIC} and performs weighted $K$-nearest-neighbor (WKNN) matching against the fingerprint database. \emph{PCA--WKNN} replaces the model-based feature extractor with principal component analysis (PCA)~\cite{Jolliffe2002PCA} before applying the same WKNN estimator. These baselines represent geometry-driven and data-driven fingerprint compression for localization, respectively.

	For fairness, all three WFM baselines are pretrained from scratch using the
	same DeepMIMO corpus rather than initialized with officially released weights. The
	physical CSI samples and splits are shared, while only the input rearrangement is
	adapted to each native tokenizer. It is worth noting that LWM~v1.1 and ContraWiMAE perform masked modeling only in the spatial--frequency domain. Accordingly, WGPT retains all 14 OFDM symbols and uses tensors of size $(N,16,N_{\mathrm{sc}},14)$, whereas the other two treat the temporal dimension as independent samples and use $(14N,16,N_{\mathrm{sc}})$. 
	In our experiments, the selected layers always
	correspond to the first, central, and last encoder blocks: $\{1,5,8\}$ for the
	eight-layer WGPT and $\{1,6,12\}$ for the twelve-layer backbones. Thus, RAFC
	receives comparable low-, mid-, and high-level representations without forcing
	the backbones to share an artificial input layout. Within each task, paired methods
	use identical seeds, heads, splits, and optimization budgets; statistical
	significance is assessed using a two-sided paired $t$-test and reported as a $p$-value.
	
	\subsection{Performance on Simulation Benchmarks}
	
	\begin{table*}[t] 
		\centering
		\caption{Overall performance across downstream tasks.
			Results are reported as mean $\pm$ standard deviation over five paired
			random seeds. The $p$-values are obtained using two-sided paired $t$-tests
			between each WFM and its RAFC-enhanced counterpart.}
		\label{tab:overall_rafc_results}
		\resizebox{\textwidth}{!}{%
			\begin{tabular}{lccc@{\hspace{0.8em}}ccc@{\hspace{0.8em}}ccc}
				\toprule
				\multirow{2}{*}{\textbf{Task}}
				& \multicolumn{3}{c}{\textbf{WGPT} \cite{WirelessGPT}}
				& \multicolumn{3}{c}{\textbf{LWM v1.1} \cite{LWM}}
				& \multicolumn{3}{c}{\textbf{ContraWiMAE} \cite{ContraWiMAE}} \\
				\cmidrule(lr){2-4}
				\cmidrule(lr){5-7}
				\cmidrule(lr){8-10}
				& \textbf{Base}
				& \textbf{+RAFC}
				& $\boldsymbol{p}$
				& \textbf{Base}
				& \textbf{+RAFC}
				& $\boldsymbol{p}$
				& \textbf{Base}
				& \textbf{+RAFC}
				& $\boldsymbol{p}$ \\
				\midrule
				
				\makecell[l]{Channel estimation\\
					($\overline{\mathrm{NMSE}}$ (dB) $\downarrow$)}
				& $-17.21 \pm 0.13$
				& $\mathbf{-18.82 \pm 0.05}$
				& $<0.001$
				& $-16.29\pm 0.15$ & $\mathbf{-17.84\pm 0.09}$ & $<0.001$
				& $-15.68\pm 0.08$ & $\mathbf{-16.59\pm 0.07}$ & $<0.001$ \\
				
				\midrule
				\makecell[l]{Channel prediction\\
					($\overline{\mathrm{NMSE}}$ (dB) $\downarrow$)}
				& $-16.42 \pm 1.21$
				& $\mathbf{-17.86 \pm 0.98}$
				& $0.094$
				& $-15.58\pm 0.84$ & $\mathbf{-16.60\pm 0.69}$ & $0.084$
				& $-14.56\pm 1.36$ & $\mathbf{-15.07\pm 0.73}$ & $0.132$ \\
				
				\midrule
				\makecell[l]{Beam prediction\\
					(64-entry codebook)\\
					($\overline{\mathrm{Top\text{-}1}}\uparrow$, \%)}
				& $64.56 \pm 0.47$
				& $\mathbf{66.44 \pm 0.71}$
				& $0.017$
				& $62.79 \pm 0.61$
				& $\mathbf{63.40 \pm 0.87}$
				& $0.342$
				& $41.49 \pm 0.82$
				& $\mathbf{42.30 \pm 0.60}$
				& $0.045$ \\
				
				\midrule
				\makecell[l]{Localization\\
					($\overline{\mathrm{MEE}}\downarrow$)}
				& $6.38 \pm 0.40$
				& $\mathbf{4.09 \pm 0.37}$
				& $0.002$
				& $7.96 \pm 1.12$
				& $\mathbf{4.94 \pm 0.81}$
				& $<0.001$
				& $6.41 \pm 0.51$
				& $\mathbf{5.24 \pm 0.33}$
				& $0.003$ \\
				
				\bottomrule
			\end{tabular}%
		}
	\end{table*}

	\begin{figure*}[t] 
		\centering
		{
			\includegraphics[width=0.7\textwidth]{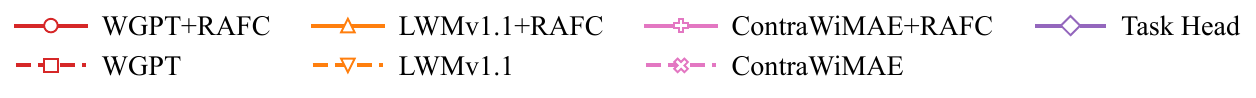}
		}\\
		\begin{subfigure}{0.32\linewidth}
			\centering
			\includegraphics[width=\linewidth]{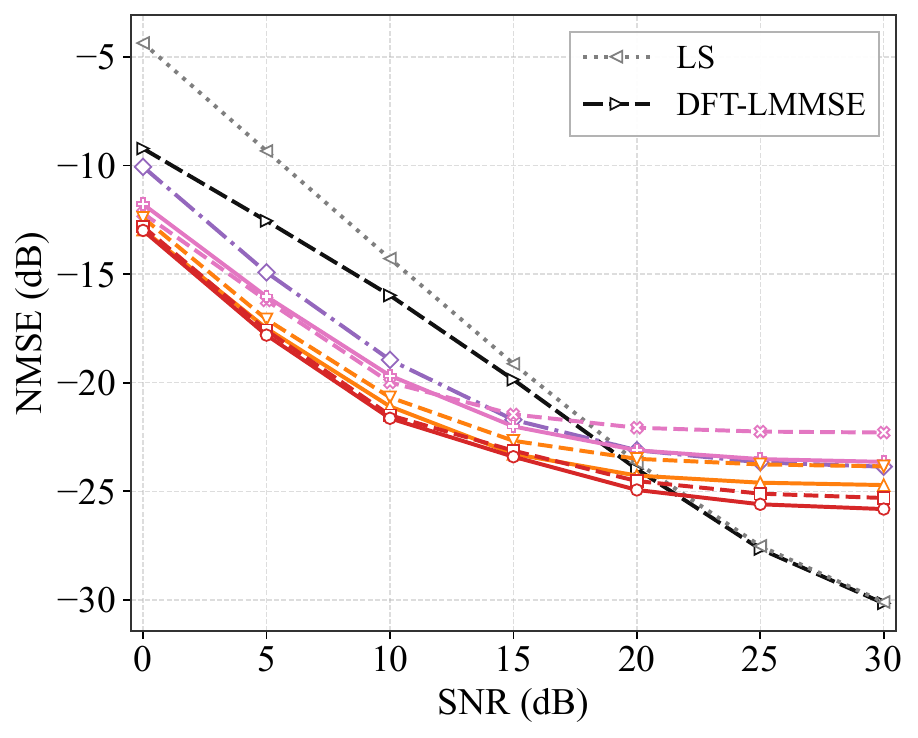}
			\caption{}
			\label{fig:est}
		\end{subfigure}
		\begin{subfigure}{0.32\linewidth}
			\centering
			\includegraphics[width=\linewidth]{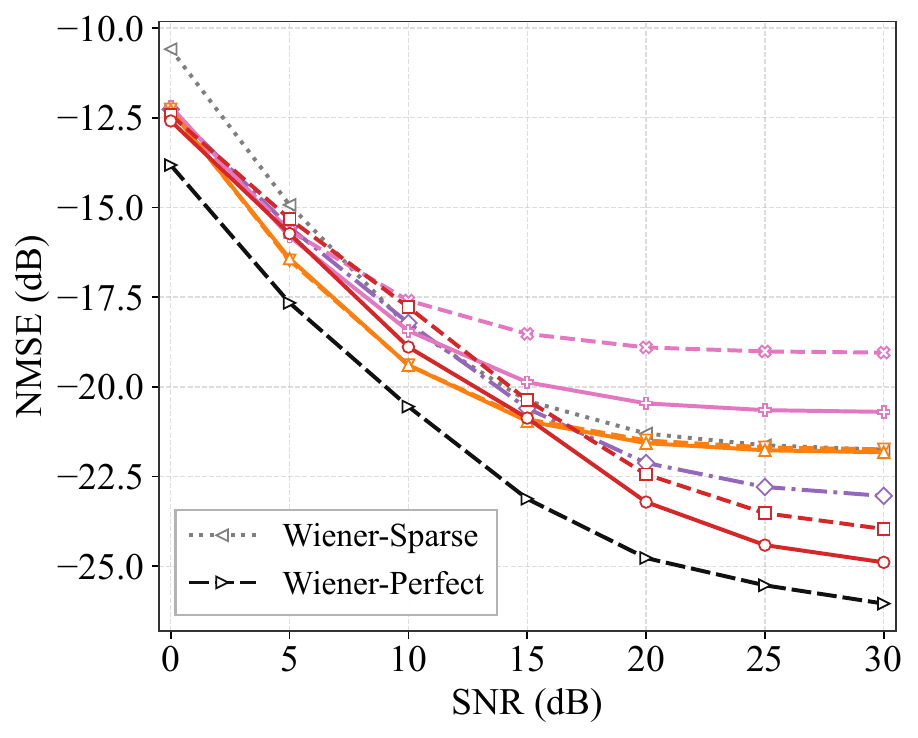}
			\caption{}
			\label{fig:pred}
		\end{subfigure}
		\begin{subfigure}{0.33\linewidth}
			\centering
			\includegraphics[width=\linewidth]{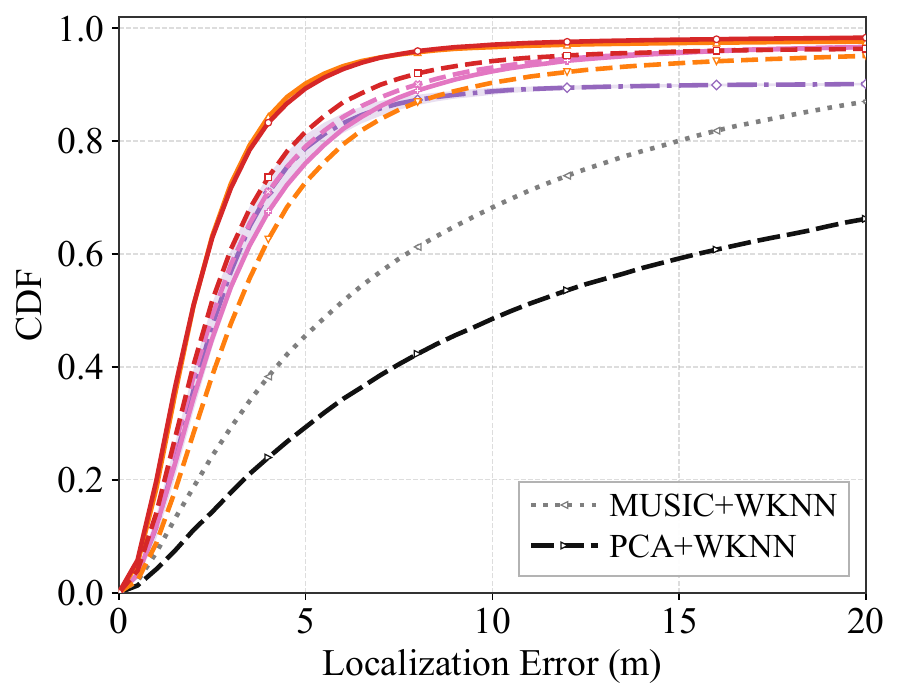}
			\caption{}
			\label{fig:loc}
		\end{subfigure}
		\captionsetup{font={small}, justification=raggedright}
		\caption{Task-wise performance: (a) channel-estimation NMSE;
			(b) channel-prediction NMSE; and (c) localization-error CDF.}
		\label{fig:all_results}
	\end{figure*}
	
	This subsection addresses two complementary questions. First, we test whether
	the benefit of RAFC is preserved across WFM architectures, which isolates the
	generality of the adaptation interface from the strength of a particular
	backbone. Second, we examine each downstream task against its specialized
	baselines to identify when multi-depth features help and what physical
	information accounts for the gain.

	\subsubsection{Backbone-Wise Gains of RAFC}
	Table~\ref{tab:overall_rafc_results} evaluates RAFC with three frozen WFM
	backbones that differ in depth, tokenization, and hidden dimension. RAFC improves
	the numerical mean for all 12 reported backbone--task pairs, although the gain and
	its statistical significance depend on the task. For channel estimation, the
	SNR-averaged NMSE improves by 1.61, 1.55, and 0.91~dB for WGPT, LWM~v1.1,
	and ContraWiMAE, respectively, with $p=0.001$ in all three cases. Channel
	prediction shows gains of 1.44, 1.02, and 0.51~dB, but their $p$-values of
	0.094, 0.084, and 0.132 do not reach the 0.05 significance level. For beam
	prediction, RAFC raises the SNR-averaged Top-1 accuracy by 1.88, 0.61, and
	0.81 percentage points; the gains are significant for WGPT and ContraWiMAE but
	not for LWM~v1.1. The largest and most consistent improvements occur in
	localization, where RAFC reduces MEE from 6.38 to 4.09~m, from 7.96 to
	4.94~m, and from 6.41 to 5.24~m, with all three comparisons significant.

	These results show that RAFC is a reusable adaptation interface rather than a
	backbone-specific modification, but they also clarify where multi-depth composition
	is most effective. Channel estimation and localization benefit reliably because
	they require local channel or geometric detail together with broader contextual
	information. Beam prediction relies more strongly on high-level spatial context,
	so the final-layer representation is already competitive and leaves less room for
	routing gains. Channel prediction is the most variable task: RAFC improves all
	three means, but temporal extrapolation remains sensitive to seed-dependent
	optimization and to the temporal information retained by each backbone. Thus,
	RAFC can reorganize complementary representations across depth, but cannot recover
	information that the frozen backbone has not preserved.

	\subsubsection{Task-Wise Comparison with Specialized Baselines}
	Next, we provide detailed comparisons between RAFC and conventional baselines across the four downstream tasks.

	\textit{Channel estimation:}
	The input is the noisy LS estimate $\widetilde{\bm{H}}_t$ observed on a fixed 5G New Radio (NR) Type-I demodulation reference signal (DMRS) mask. Pilots are spaced every four subcarriers in frequency and placed at OFDM symbols 0 and 11 in time over a resource grid of 72 subcarriers and 14 OFDM symbols. The target is the complete current-slot channel $\bm{H}_t$. As shown in Fig.~\ref{fig:est}, WGPT+RAFC improves the final-layer
	WGPT baseline throughout the SNR range and achieves the best learned-model result
	at every operating point. The LWM~v1.1- and ContraWiMAE-based variants show a
	similar RAFC benefit mainly at moderate and high SNRs, where representation quality
	rather than pilot noise becomes the dominant limitation. Below approximately
	20~dB, the learned models outperform LS and DFT-LMMSE by exploiting channel priors
	to suppress noise and interpolate missing resource elements. Beyond 20~dB, however,
	the learned curves approach model-dependent floors, whereas the two conventional
	estimators continue to improve and overtake them between 20 and 25~dB. RAFC lowers
	the learned-model floor by combining pilot-local amplitude and phase details from
	shallow layers with the broader channel context retained at later depths.

	\begin{table*}[t] 	
		\centering
		\caption{Beam prediction performance under codebook sizes 32 and 64.
			Results are reported as mean $\pm$ standard deviation over five random seeds.}
		\label{tab:beam_prediction_results}
		\resizebox{\textwidth}{!}{%
			\begin{tabular}{lccccccc}
				\toprule
				\textbf{Metric}
				& \textbf{WGPT+RAFC}
				& \textbf{WGPT}
				& \textbf{LWM v1.1+RAFC}
				& \textbf{LWM v1.1}
				& \textbf{ContraWiMAE+RAFC}
				& \textbf{ContraWiMAE}
				& \textbf{Task Head} \\
				\midrule
				
				\multicolumn{8}{c}{\textbf{Codebook size = 32}} \\
				\midrule
				Top-1 Accuracy (\%)
				& $80.72 \pm 0.58$
				& $80.06 \pm 0.41$
				& $78.92 \pm 0.46$
				& $78.60 \pm 0.74$
				& $64.32 \pm 1.60$
				& $63.59 \pm 1.28$
				& $62.31 \pm 0.94$ \\
				
				Macro-F1 (\%)
				& $74.32 \pm 0.72$
				& $73.31 \pm 0.52$
				& $73.60 \pm 0.65$
				& $73.14 \pm 1.18$
				& $56.14 \pm 2.48$
				& $53.33 \pm 1.99$
				& $50.69 \pm 1.38$ \\
				
				Weighted-F1 (\%)
				& $80.59 \pm 0.54$
				& $79.98 \pm 0.43$
				& $78.81 \pm 0.46$
				& $78.50 \pm 0.66$
				& $64.08 \pm 1.75$
				& $61.47 \pm 1.34$
				& $62.03 \pm 1.01$ \\
				
				\midrule
				\multicolumn{8}{c}{\textbf{Codebook size = 64}} \\
				\midrule
				Top-1 Accuracy (\%)
				& $66.44 \pm 0.71$
				& $64.56 \pm 0.47$
				& $63.40 \pm 0.87$
				& $62.79 \pm 0.61$
				& $42.30 \pm 0.60$
				& $41.49 \pm 0.82$
				& $43.11 \pm 1.05$ \\
				
				Macro-F1 (\%)
				& $55.61 \pm 0.94$
				& $53.75 \pm 0.75$
				& $54.38 \pm 1.30$
				& $53.22 \pm 1.19$
				& $31.24 \pm 0.85$
				& $29.63 \pm 0.78$
				& $30.42 \pm 0.86$ \\
				
				Weighted-F1 (\%)
				& $66.23 \pm 0.67$
				& $64.13 \pm 0.39$
				& $63.09 \pm 0.89$
				& $62.46 \pm 0.62$
				& $41.83 \pm 0.68$
				& $39.47 \pm 0.70$
				& $42.49 \pm 1.10$ \\
				
				\bottomrule
			\end{tabular}%
		}
	\end{table*}
	\begin{table}[t]	
		\centering
		\caption{Localization performance of different schemes. Results are
			reported as the mean Euclidean error (MEE), mean $\pm$ standard
			deviation, and the corresponding 95\% confidence interval (CI) over five
			random seeds.}
		\label{tab:localization_results}
		\setlength{\tabcolsep}{5pt}
		\renewcommand{\arraystretch}{1.1}
		\footnotesize
		\begin{tabular}{lcc}
			\toprule
			\textbf{Scheme}
			& \textbf{MEE (Mean $\boldsymbol{\pm}$ Std)}
			& \textbf{95\% CI} \\
			\midrule
			WGPT+RAFC
			& $\mathbf{4.094 \pm 0.371}$ m
			& $[3.633,\,4.555]$ \\
			WGPT
			& $6.377 \pm 0.401$ m
			& $[5.879,\,6.876]$ \\
			LWM v1.1+RAFC
			& $4.938 \pm 0.811$ m
			& $[3.932,\,5.945]$ \\
			LWM v1.1
			& $7.961 \pm 1.118$ m
			& $[6.573,\,9.349]$ \\
			ContraWiMAE+RAFC
			& $5.241 \pm 0.332$ m
			& $[4.829, 5.653]$ \\
			ContraWiMAE
			& $6.409 \pm 0.512$ m
			& $[5.773, 7.045]$ \\
			Task Head
			& $4.767 \pm 0.295$ m
			& $[4.401,\,5.134]$ \\
			MUSIC+WKNN
			& $10.026 \pm 0.156$ m
			& $[9.832,\,10.220]$ \\
			PCA+WKNN
			& $22.758 \pm 0.299$ m
			& $[22.387,\,23.129]$ \\
			\bottomrule
		\end{tabular}
	\end{table}

	\textit{Channel prediction:}
	Here the same sparse current-slot observation $\widetilde{\bm{H}}_t$ is used to
	predict the complete next-slot channel $\bm{H}_{t+1}$. The problem therefore
	contains two coupled uncertainties---completion of the current channel and
	extrapolation of its temporal evolution. Fig.~\ref{fig:pred} shows that
	Wiener-Perfect remains the upper-bound reference throughout the SNR range because
	it receives complete current-slot CSI. Among methods using sparse CSI, the strongest
	learned variants outperform Wiener-Sparse. At low SNR, their differences are
	small because all methods are observation-noise limited. At moderate SNRs,
	LWM~v1.1+RAFC is competitive, while above 15~dB WGPT+RAFC continues to improve
	and its advantage over final-layer WGPT widens. In contrast, the LWM~v1.1- and
	ContraWiMAE-based curves saturate earlier because these backbones model each OFDM
	symbol in the spatial--frequency domain without explicitly encoding inter-symbol
	evolution. The remaining gap to Wiener-Perfect indicates that sparse observation
	still limits prediction, while the RAFC gains at moderate and high SNRs show that
	multi-depth features preserve additional phase-progression and path-evolution cues.

	\textit{Beam prediction:}
	The model receives $\widetilde{\bm{H}}_t$ and predicts the next-slot codeword $c_{t+1}^{\star}$ that maximizes the received power in Eq.~\eqref{eq:optimal_beam}. In general, beam prediction relies more strongly on high-level spatial representations than the other downstream tasks, making the final-layer feature already relatively effective and leaving limited room for RAFC improvement. Table~\ref{tab:beam_prediction_results} compares the beam-prediction performance of the three backbone families under 32- and 64-entry codebooks. Top-1 accuracy and the two F1 scores exhibit broadly consistent trends. RAFC improves the performance of all three backbones, although the gains remain modest. When the classification problem becomes easier (the codebook size decreases), the remaining benefit of multi-level composition is further reduced. Nevertheless, the substantial advantage of the WGPT- and LWM-based models over the standalone Task Head demonstrates the value of pretrained WFM representations for beam prediction.

	\textit{Localization:}
	Unlike the preceding tasks, localization is evaluated in the unseen City~16 scenario, which covers an area of approximately \(207\times200~\mathrm{m}^2\). Each input consists of the full CSI tensor \(\bm{H}_t\) collected over 144 subcarriers with $32\times2$ antennas, and the target is the two-dimensional UE position \(\bm{p}_t\). As shown in Fig.~\ref{fig:loc} and
	Table~\ref{tab:localization_results}, WGPT+RAFC attains the lowest MEE of
	$4.094$~m and produces one of the steepest localization-error CDFs. LWM~v1.1+RAFC is
	competitive over the central-error region, while ContraWiMAE+RAFC provides a
	smaller but still visible shift relative to its final-layer counterpart. Relative
	to the three final-layer baselines, RAFC reduces MEE by 35.8\%, 38.0\%, and
	18.2\%, respectively. The consistent movement of the CDFs toward smaller errors
	shows that these gains extend across the test distribution rather than arising
	from a small subset of favorable positions.
	\begin{figure}[t]
		\centering
		\includegraphics[width=0.4\textwidth]{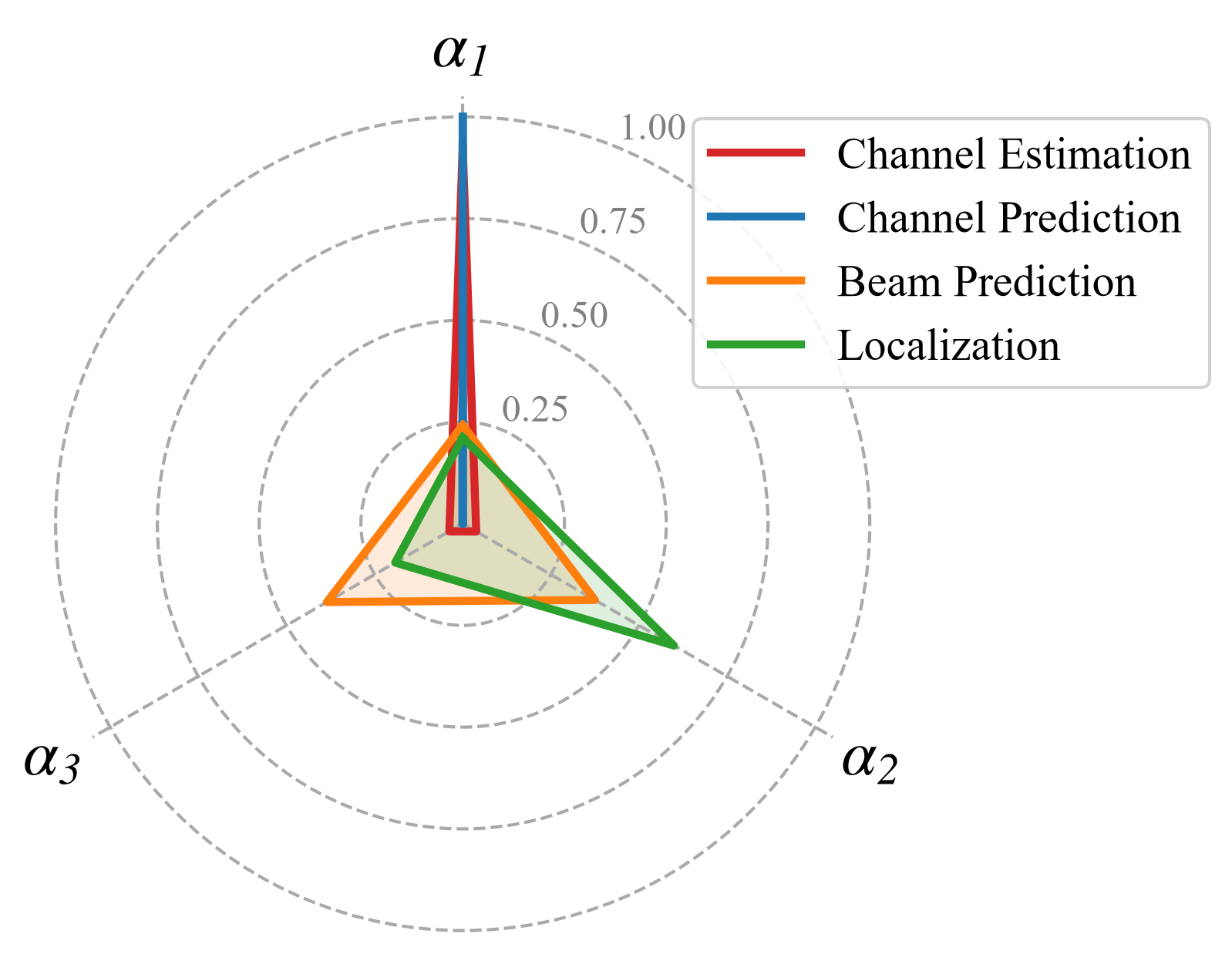}
		\captionsetup{font={small}, justification=raggedright}
		\caption{Converged task-specific routing profiles over three WFM layers.}
		\label{fig:task_routing_radar}
	\end{figure}

	The comparison also explains why multi-depth composition is useful
	for localization. MUSIC--WKNN retains an explicit angular cue but cannot fully
	resolve environment-dependent multipath ambiguity; PCA--WKNN preserves dominant
	variance, which need not coincide with position-discriminative variation. A
	standalone task head remains competitive with several frozen-WFM variants,
	exposing representation mismatch when an overly invariant final layer is used
	without adaptation. RAFC removes this bottleneck:
	intermediate features suppress nuisance fading while retaining spatial relations,
	and shallow and deep features contribute fine geometry and scene-level context. Its
	improvement over both the task head and the model-based fingerprints is therefore
	evidence of better representation alignment, rather than simply a more powerful
	regressor.
	
			\begin{figure*}[t]
		\centering
		
		\includegraphics[width=0.34\textwidth]{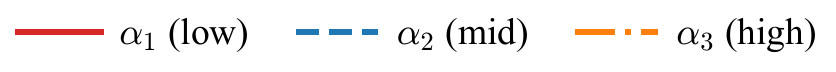}
		\\
		\centering
		\centering
		\includegraphics[width=\linewidth]{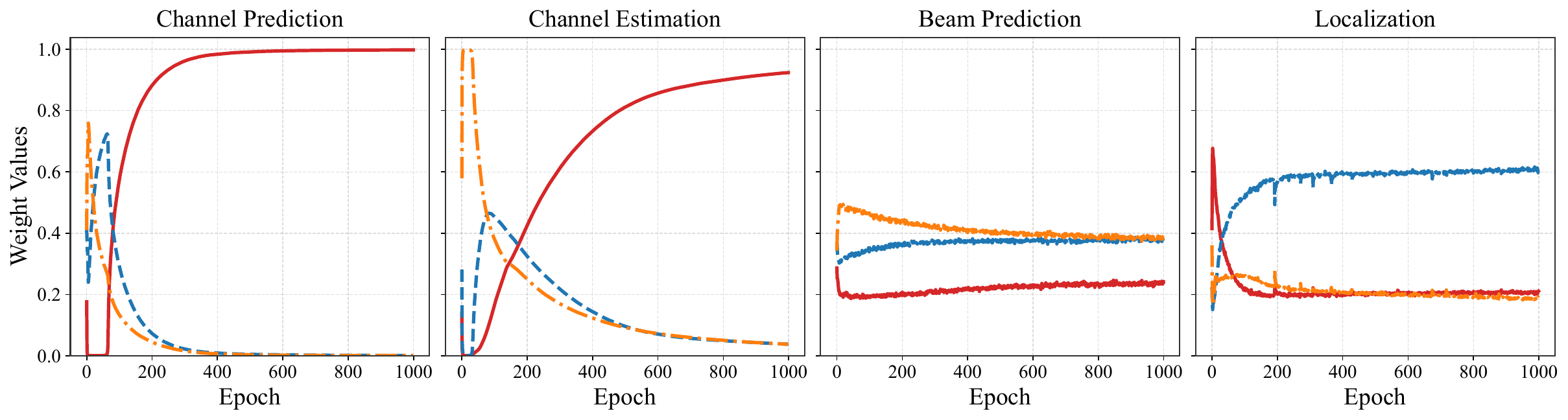}
		\captionsetup{font={small}, justification=raggedright}
		\caption{Evolution of the RAFC routing weights during training for (a) channel prediction, (b) channel estimation, (c) beam prediction, and (d) localization.}
		\label{fig:weights_propagation}
	\end{figure*}
	
	\begin{figure*}[t]
		\centering
		{
			\includegraphics[width=1\textwidth]{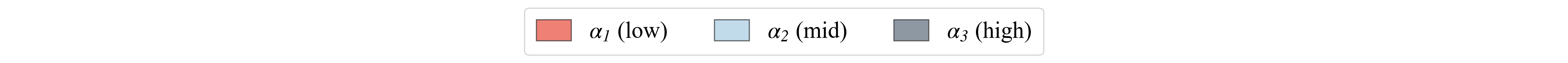}
		}\\
		\begin{subfigure}{0.4\textwidth}
			\centering
			\includegraphics[width=\linewidth]{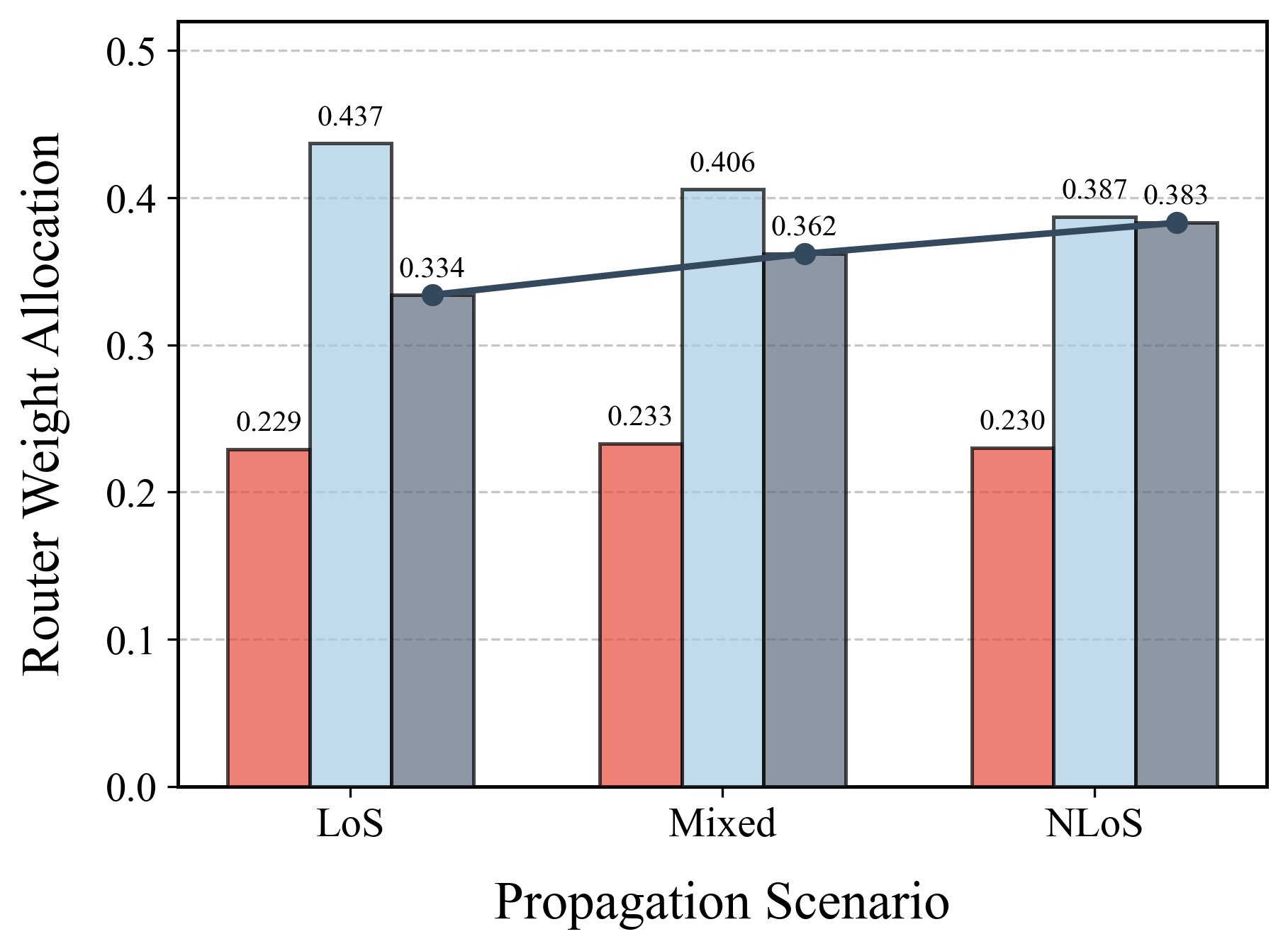}
			\caption{}
			\label{fig:weight_los_nlos}
		\end{subfigure}
		\qquad\qquad
		\begin{subfigure}{0.4\textwidth}
			\centering
			\includegraphics[width=\linewidth]{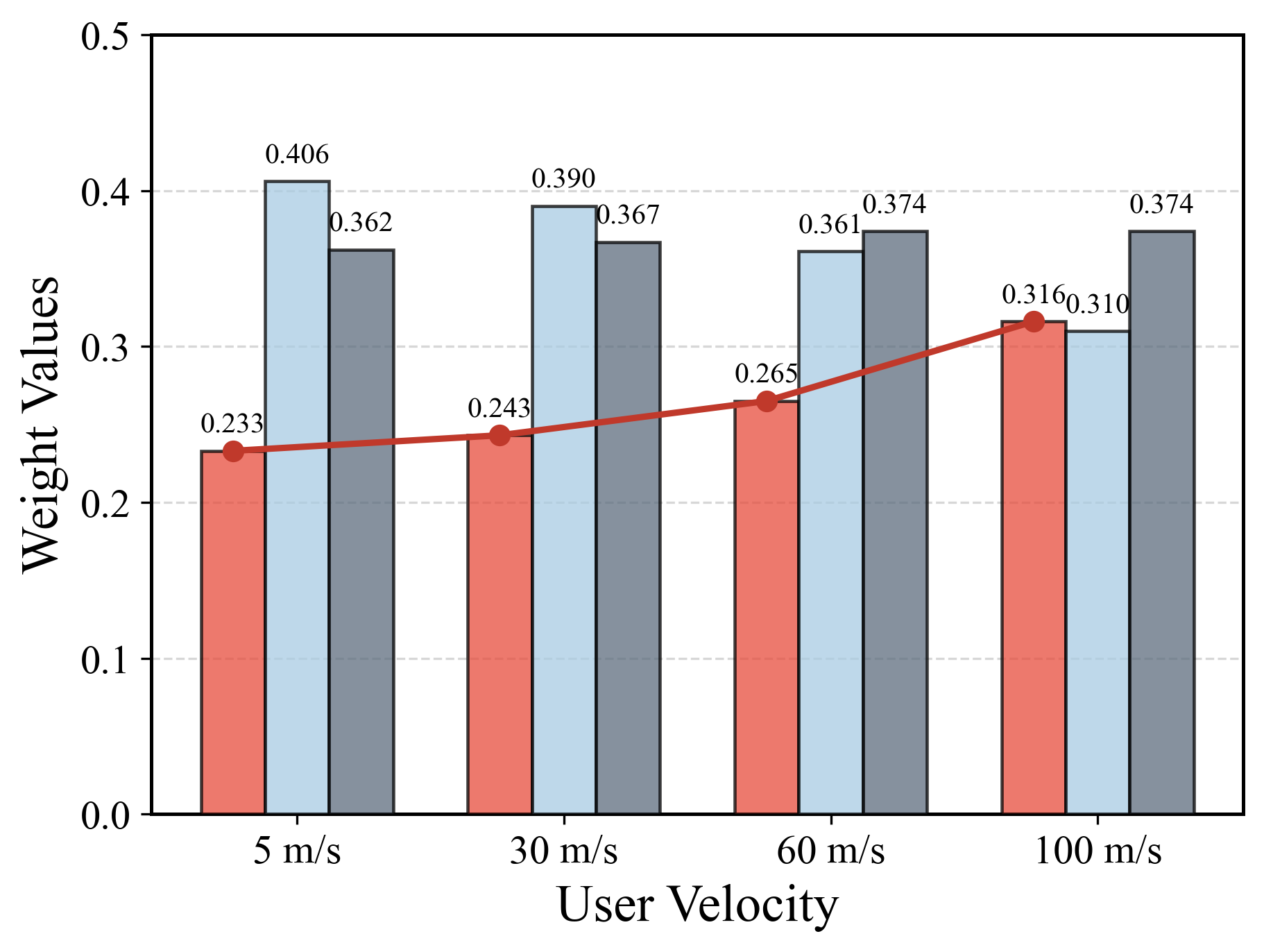}
			\caption{}
			\label{fig:weight_velocity}
		\end{subfigure}
		\captionsetup{font={small}, justification=raggedright}
		\caption{Sample-condition adaptation of RAFC for beam prediction:
			(a) line-of-sight (LoS), mixed, and non-line-of-sight (NLoS) propagation; and (b) terminal velocity.}
		\label{fig:scenarios_adaptation}
	\end{figure*}
	
	\subsection{Mechanism, Ablation, and Efficiency Analysis of RAFC}

	The preceding results establish the performance benefit, but do not by themselves
	show whether the router learns physically meaningful composition policies. We
	therefore examine the learned weights at both task and sample levels, and then
	ablate adaptation depth, layer selection, and pooling.
	
	\subsubsection{Task- and Sample-Adaptive Routing}

	The converged profiles in
	Fig.~\ref{fig:task_routing_radar} separate the tasks according to the
	physical information in their targets. Channel estimation and prediction assign
	nearly all mass to the shallow layer because their losses are evaluated directly
	on complex-valued channel coefficients; small local phase or amplitude errors
	cannot be replaced by a semantically invariant representation. Their small
	nonzero contributions from later layers act as contextual regularization, while the
	router limits this context from overwriting signal details. Beam prediction and localization instead require invariance to channel
	perturbations that do not change the optimal beam or position. Beam prediction
	distributes its mass across all three depths: shallow phase cues distinguish
	adjacent beams, while intermediate/deep features integrate them into angular and
	multipath context. Localization places the largest mass on the intermediate layer,
	which provides a useful compromise between noisy local fading and an overly
	compressed final representation. These profiles supply a mechanistic explanation
	for the task-wise gains in Section~IV-B: WFM depth forms a progression from
	measurement fidelity to invariance, and each target has a different optimal point
	along that progression.
	
	Fig.~\ref{fig:weights_propagation} then traces the learned routing weights for the four tasks at each training epoch, showing that they undergo nonmonotonic transitions before converging to stable task-specific profiles. This behavior is particularly evident in channel estimation and prediction, where RAFC initially assigns substantial weight to intermediate and deep features to establish the global channel structure required for completion or extrapolation. As this coarse structure is learned, the router gradually shifts toward shallow features to refine local phase and amplitude details, eventually producing a shallow-layer-dominated representation.

	Task identity is not the only factor. As shown in
	Fig.~\ref{fig:scenarios_adaptation}, the beam-prediction router changes its policy
	with the propagation state of individual samples. From LoS to NLoS, the
	deep-layer weight increases from 0.334 to 0.383 while the intermediate-layer
	weight decreases from 0.437 to 0.387. Under LoS, a relatively coherent angular
	structure can be resolved from mid-level features; under NLoS, individual path
	phases become less stable and the beam decision depends more on globally
	aggregated multipath context, explaining the shift toward the deep layer. A different reallocation occurs with mobility. As velocity rises from 5 to
	100~m/s, the shallow-layer weight increases from 0.233 to 0.316 and the
	intermediate-layer weight decreases from 0.406 to 0.310. Rapid temporal variation
	makes instantaneous phase progression more informative and reduces the validity
	of slowly varying abstractions, so the router returns weight to local features.
	The opposite responses to multipath complexity and mobility are especially
	informative: RAFC is not learning a fixed ``hard-sample'' rule, but is selecting
	the abstraction level that matches the physical source of uncertainty.
	
	\begin{table}[t]
		\centering
		\caption{Ablation study on WGPT at a fixed inference SNR of 15~dB. Results are reported as mean $\pm$ standard deviation over five random seeds. Extra FLOPs denote the sum of forward- and backward-propagation FLOPs relative to the frozen-WGPT baseline, excluding the FLOPs incurred by optimizer updates.}
		\label{tab:ablation_all}
		\setlength{\tabcolsep}{3pt}
		\renewcommand{\arraystretch}{1.08}
		\footnotesize
		
		\resizebox{\columnwidth}{!}{%
			\begin{tabular}{lccc}
				\toprule
				\textbf{Method}
				& \makecell[c]{\textbf{Beam Pred.}\\\textbf{Acc. (\%)}}
				& \makecell[c]{\textbf{Channel Pred.}\\\textbf{NMSE (dB)}}
				& \makecell[c]{\textbf{Extra FLOPs},\\ \textbf{M/sample}} \\
				\midrule
				
				\rowcolor{gray!20}
				\multicolumn{4}{c}{\textbf{Fine-Tuning Strategies}} \\
				
				WGPT
				& $67.61\pm0.59$
				& $-20.37\pm 1.06$
				& $0$ \\
				
				WGPT-RAFC
				& $71.83\pm1.17$
				& $-20.87\pm1.21$
				& $0.296$ \\
				
				WGPT-BMFT
				& $68.33\pm1.17$
				& $-16.95\pm1.61$
				& $390.760$ \\
				
				WGPT-Full-FT
				& $\mathbf{75.35\pm0.81}$
				& $\mathbf{-24.67\pm0.41}$
				& \makecell[c]{$4750.304$} \\
				
				\addlinespace[3pt]
				
				\rowcolor{gray!20}
				\multicolumn{4}{c}{\textbf{Layer Selection Strategies}} \\
				
				WGPT
				& $67.61\pm0.59$
				& $-20.37\pm 1.06$
				& $0$ \\
				
				\makecell[l]{WGPT-RAFC(1,5,8)}
				& $71.83\pm1.17$
				& $\mathbf{-20.87\pm1.21}$
				& $0.296$ \\
				
				(1,2,3)
				& $62.48\pm2.24$
				& $-20.30\pm0.41$
				& $0.296$ \\
				
				(6,7,8)
				& $70.78\pm0.58$
				& $-19.51\pm0.23$
				& $0.296$ \\
				
				(1,2,$\cdots$,8)
				& $\mathbf{72.72\pm1.85}$
				& $-20.81\pm1.69$
				& $0.790$ \\
				
				\addlinespace[3pt]
				
				\rowcolor{gray!20}
				\multicolumn{4}{c}{\textbf{Pooling Strategies}} \\
				
				WGPT
				& $67.61\pm0.59$
				& $-20.37\pm 1.06$
				& $0$ \\
				
				\makecell[l]{WGPT-RAFC\\(Mean pooling)}
				& $71.83\pm1.17$
				& $-20.87\pm1.21$
				& $0.296$ \\
				
				Max pooling
				& $66.22\pm0.57$
				& $-20.61\pm1.55$
				& $0.296$\\
				
				Attention pooling
				& $69.61\pm1.32$
				& $-20.50\pm1.46$
				& $1.070$ \\
				
				Full pooling
				& $\mathbf{72.22\pm0.74}$
				& $\mathbf{-21.08\pm1.27}$
				& $49.739$ \\
				\addlinespace[3pt]
				
				\rowcolor{gray!20}
				\multicolumn{4}{c}{\textbf{Weight-Routing Strategies}} \\
				\midrule
				WGPT    & $67.61\pm0.59$
				& $-20.37\pm 1.06$
				& $0$ \\
				\makecell[l]{WGPT-RAFC\\(Dynamic routing)} 	& $\mathbf{71.83\pm1.17}$
				& $\mathbf{-20.87\pm1.21}$
				& $0.296$ \\
				Static routing & $69.39\pm 1.83$           & $-20.08\pm 1.97$ & $0.296$\\
				Uniform routing   & $65.42\pm1.32$ & $-16.87\pm 1.15$        & $0$ \\
				\bottomrule
			\end{tabular}%
		}
	\end{table}
	
	\subsubsection{Ablation Studies and Complexity Analysis}
	Using WGPT as the backbone, we conduct ablation studies on four components of RAFC: the adaptation strategy, layer selection, pooling-based feature compression, and weight routing. We also report the additional computational cost of each variant in floating-point operations (FLOPs) per sample. For adaptation, RAFC is compared with the frozen WGPT baseline, best-matched fine-tuning (WGPT-BMFT\footnote{In the BMFT implementation, the final MLP sublayer of WGPT is unfrozen, resulting in 0.0504M trainable parameters, comparable to the 0.049M parameters introduced by RAFC.}), which updates a parameter-matched subset of the backbone, and full fine-tuning (WGPT-Full-FT), which updates the entire backbone. For layer selection, the proposed spaced configuration $(1,5,8)$ is compared with consecutive shallow layers $(1,2,3)$, consecutive deep layers $(6,7,8)$, and all eight backbone layers. For feature compression, mean pooling is compared with max pooling, attention pooling, and full token-level routing without compact feature summarization. For weight routing, the proposed sample-dependent dynamic routing is compared with static routing, which learns one task-level weight vector shared by all samples, and uniform routing, which assigns equal weights to the selected layers. Beam prediction and channel prediction are selected because they exhibit distinct preferences for high-level spatial context and low-level channel details, respectively.

	\textit{Adaptation strategy:}
	At 15~dB, RAFC improves beam accuracy from 67.61\% to 71.83\% and channel-
	prediction NMSE from $-20.37$ to $-20.87$~dB, while adding only 0.296~M
	training FLOPs per sample. Full fine-tuning provides the performance upper bound, but its
	additional cost is approximately $1.6\times10^4$ times that of RAFC. BMFT is
	also substantially more expensive and degrades channel prediction. This
	degradation is consistent with partial fine-tuning disturbing a pretrained feature
	hierarchy using a comparatively small downstream set: changing only selected
	blocks can break the alignment between frozen and updated layers without providing
	the flexibility of full fine-tuning. RAFC avoids this form of representation drift
	by leaving the hierarchy intact and changing only how its existing information is
	read out. It therefore occupies a favorable performance--cost point rather than
	attempting to reproduce unconstrained full fine-tuning.
	
	\begin{table}[t]
		\caption{Inference latency for beam prediction.
			Values are reported in ms as mean (standard deviation); 0.629(19) denotes $0.629\pm0.019$~ms.}
		\label{tab:inference_latency}
		\centering
		\footnotesize
		\setlength{\tabcolsep}{2.5pt}
		\renewcommand{\arraystretch}{1.08}
		\begin{tabular*}{\columnwidth}{
				@{\extracolsep{\fill}}lcccc@{}
			}
			\toprule
			\textbf{Scheme}
			& \textbf{WFM}
			& \textbf{RAFC}
			& \textbf{Head}
			& \textbf{Total} \\
			\midrule
			WGPT
			& 0.629(19)
			& 0.047(2)
			& 0.151(4)
			& \textbf{0.828(21)} \\
			LWM~v1.1
			& 1.963(65)
			& 0.040(6)
			& 0.165(28)
			& \textbf{2.168(81)} \\
			ContraWiMAE
			& 0.535(24)
			& 0.046(3)
			& 0.138(3)
			& \textbf{0.719(29)} \\
			\bottomrule
		\end{tabular*}
		
		\vspace{1mm}
		\begin{minipage}{\columnwidth}
			\scriptsize
			\textit{Note:} Latency is measured on an NVIDIA RTX A6000 GPU
			with a batch size of one and a 32-entry beam codebook.
			Each model is warmed up for 10 iterations and subsequently timed
			over 100 runs using synchronized CUDA events under BF16 autocast.
			Inputs are preloaded onto the GPU; data loading, CPU--GPU transfer,
			and noise generation are excluded. WGPT jointly processes DMRS
			symbols 0 and 11, whereas LWM~v1.1 and ContraWiMAE use DMRS symbol
			0. WFM latency includes input normalization, tokenization, and the
			frozen-backbone forward pass. The task head is the trained two-layer
			Transformer classifier.
		\end{minipage}
		
	\end{table}
	
	\textit{Layer selection:}
	Using only shallow layers $(1,2,3)$ severely degrades beam prediction, whereas
	using only deep layers $(6,7,8)$ degrades channel prediction. The former lacks
	the receptive field needed to consolidate spatial evidence for a beam class; the
	latter has already smoothed the local phase evolution required to extrapolate the
	next channel. Exposing all eight layers raises beam accuracy by only 0.89 points
	over $(1,5,8)$, slightly worsens channel-prediction NMSE, and increases fusion
	FLOPs from 0.296 to 0.790~M. Adjacent Transformer layers are strongly correlated,
	so adding every depth mainly introduces redundant routing choices and estimation
	variance rather than new abstraction levels. The spaced $(1,5,8)$ selection acts
	as a compact basis spanning local, transitional, and global representations.
	
	\begin{figure*}[t]
		\centering
		\includegraphics[width=0.97\linewidth,trim=2 2 2 2,clip]{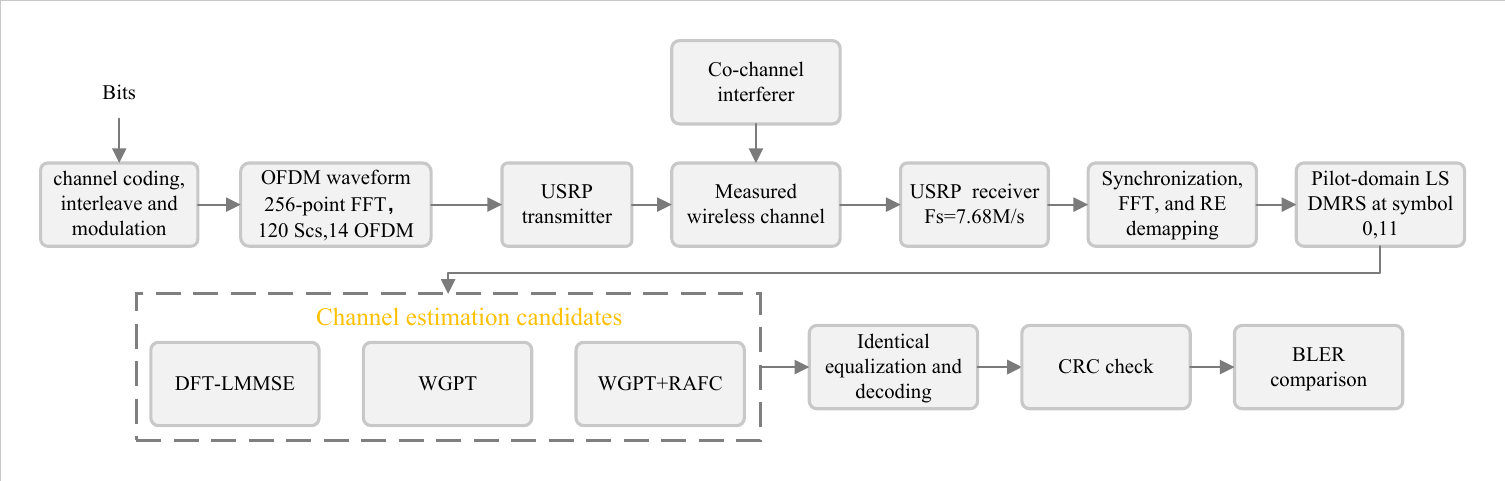}
		\caption{End-to-end evaluation workflow on the portable SDR-based mobile wireless platform.}
		\label{fig:ota_workflow}
	\end{figure*}
	
	\textit{Pooling strategy:}
	Mean pooling outperforms max and attention pooling at lower or equal cost. Full
	pooling yields small additional gains, but requires 49.739~M FLOPs, over 168
	times the cost of mean pooling. The reason is that pooling is used only to decide
	\emph{which layer} should contribute; the selected full-resolution feature maps
	are still composed after the routing weights are generated. A mean descriptor is
	therefore sufficient to summarize global channel quality and propagation regime.
	Max pooling is sensitive to isolated noisy tokens, while attention pooling adds
	capacity that can overfit the limited downstream data. The small benefit of full
	pooling confirms that routing is a low-dimensional decision even though the
	features being routed remain high dimensional.

	\textit{Weight-routing strategy:}
	Dynamic routing is the only strategy that improves both tasks over the frozen
	WGPT baseline, increasing beam-prediction accuracy by 4.22 percentage points
	and improving channel-prediction NMSE by 0.50~dB. Static routing retains a
	1.78-point beam-prediction gain but slightly degrades channel prediction, while
	uniform routing degrades both tasks. This comparison separates two sources of
	adaptivity: uniform routing ignores even the average layer preference of a task,
	whereas static routing captures that task-level preference but applies it to
	every input. Conditioning the weights on each sample allows dynamic routing to
	respond to changes in the observed channel and select a more suitable layer
	composition. Its advantage over static routing at the same reported
	0.296~M extra FLOPs therefore comes from input-dependent feature reuse.

	\textit{Complexity analysis:}
	Table~\ref{tab:inference_latency} shows that RAFC adds only
	$0.040$--$0.047$~ms, or approximately $1.8$--$6.4\%$ of the total latency,
	so inference remains dominated by the frozen WFM. LWM~v1.1 is slower despite
	its smaller parameter count because it uses 12 encoder layers rather than the
	eight used by WGPT, and its manually implemented attention and LayerNorm launch
	multiple small CUDA kernels instead of fused operators. At a batch size of one,
	these kernel-launch overheads are prominent. Thus, practical latency depends not
	only on parameter count, but also on network depth and operator fusion.

	\subsection{Portable SDR-Based Over-the-Air Validation}

	Simulation gains can still arise from regularities specific to the ray-tracing
	generator. We therefore implement the complete transceiver using portable SDR
	nodes that can be repositioned across indoor locations. The purpose is to test
	the receiver under location-dependent measured propagation, hardware
	impairments, and co-channel interference, and to determine whether the
	representation-level improvements in
	Sections~IV-B and IV-C survive after channel estimation is coupled to decoding.
	The evaluation metric is consequently changed from channel NMSE to end-to-end
	BLER.

	\begin{figure*}[t]
		\begin{subfigure}{0.60\textwidth}
			\centering
			\includegraphics[width=\linewidth]{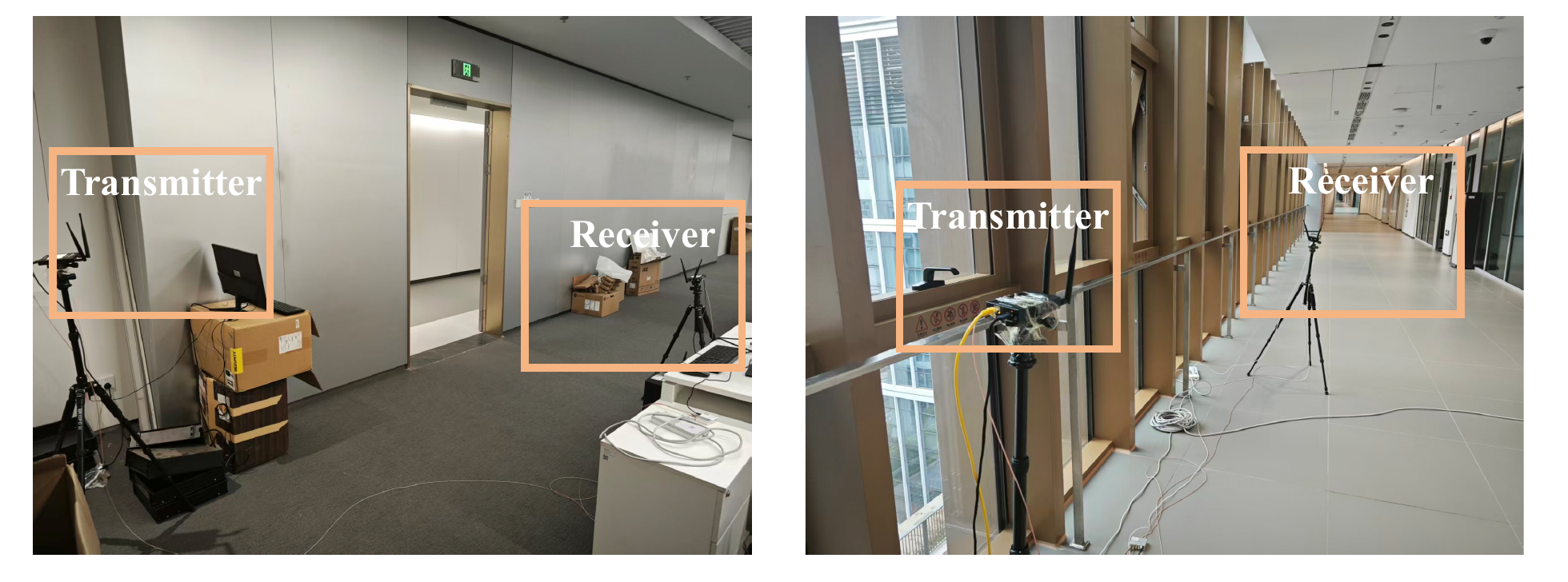}
			\caption{}
			\label{fig:data_collection}
		\end{subfigure}
		\begin{subfigure}{0.39\textwidth}
			\centering
			\includegraphics[width=\linewidth]{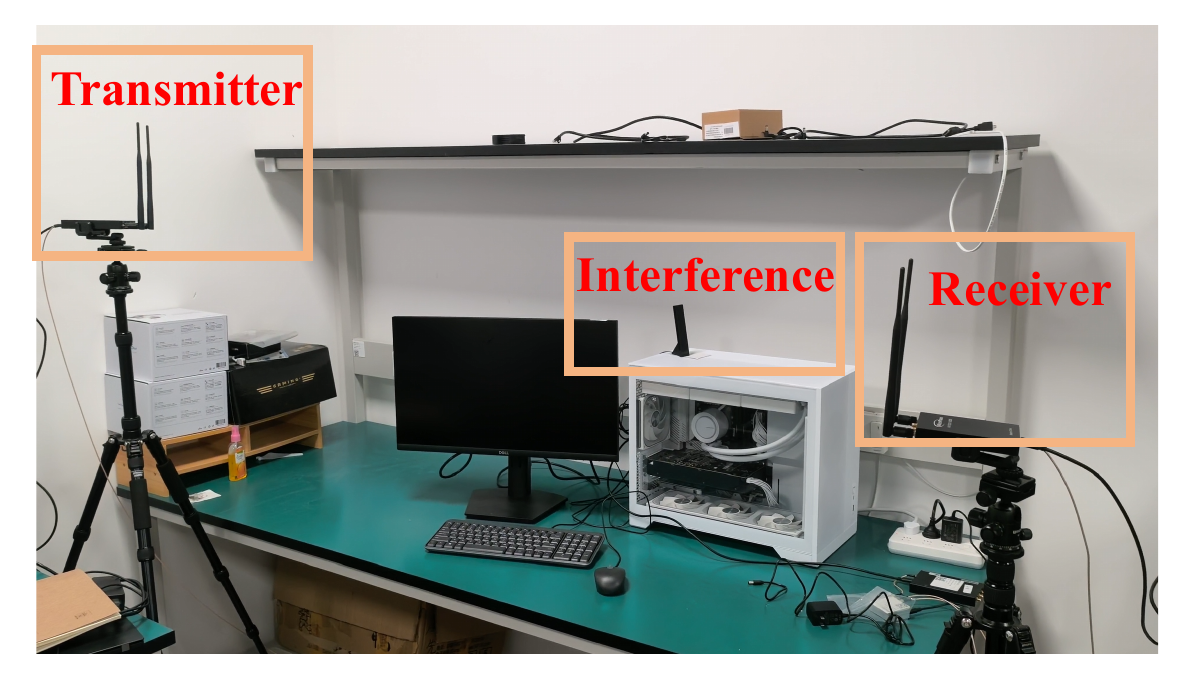}
			\caption{}
			\label{fig:datalink}
		\end{subfigure}
		\captionsetup{font={small}, justification=raggedright}
		\caption{Testing scenarios of the portable SDR-based mobile wireless platform:
			(a) the transmitter and receiver nodes are placed at different indoor office
			and corridor locations without interference; and (b) a portable co-channel
			interferer is activated in the indoor office.}
		\label{Testbed}
	\end{figure*}
	
	\begin{figure*}[t]
		\centering
		{
			\includegraphics[width=0.9\textwidth]{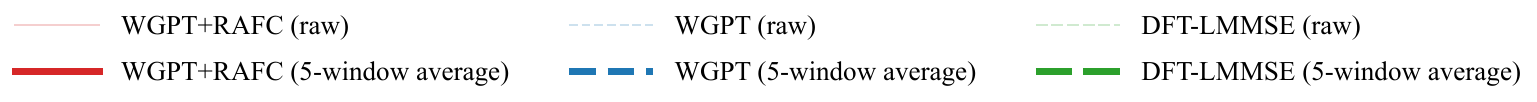}
		}\\
		\begin{subfigure}{0.4\textwidth}
			\centering
			\includegraphics[width=\linewidth]{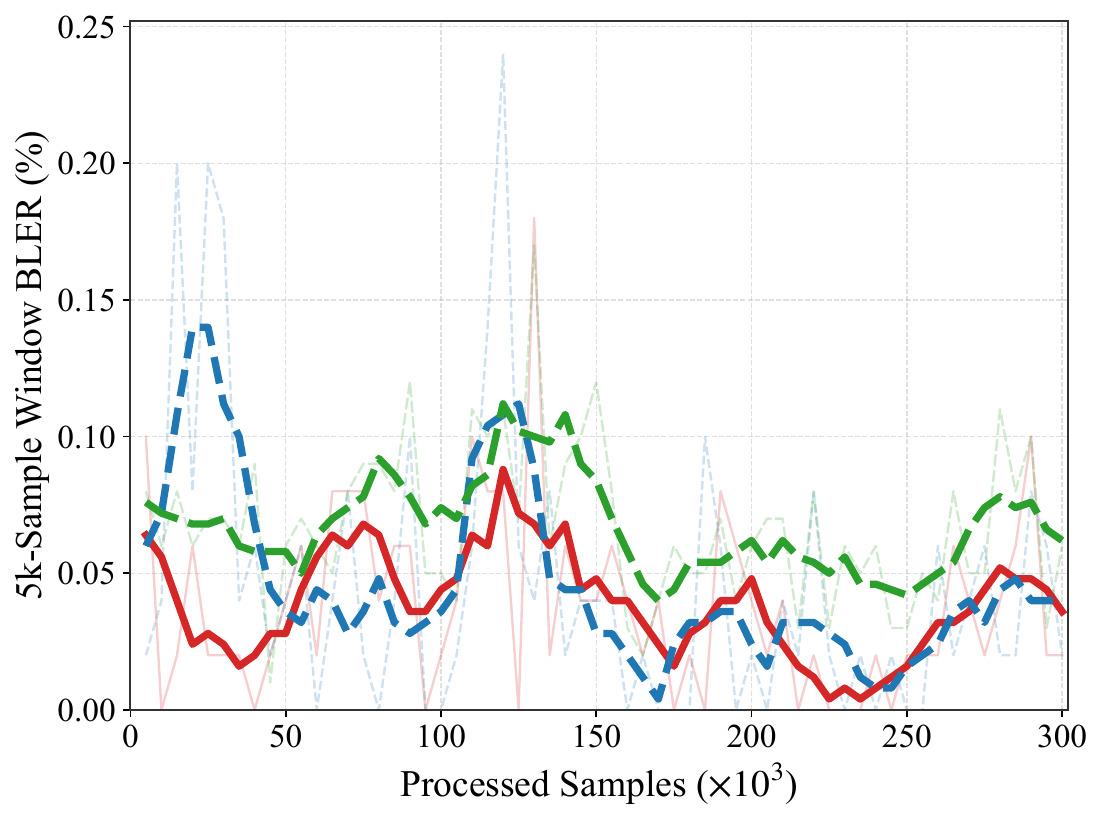}
			\caption{}
			\label{fig:without_interference}
		\end{subfigure}
		\qquad\qquad
		\begin{subfigure}{0.4\textwidth}
			\centering
			\includegraphics[width=\linewidth]{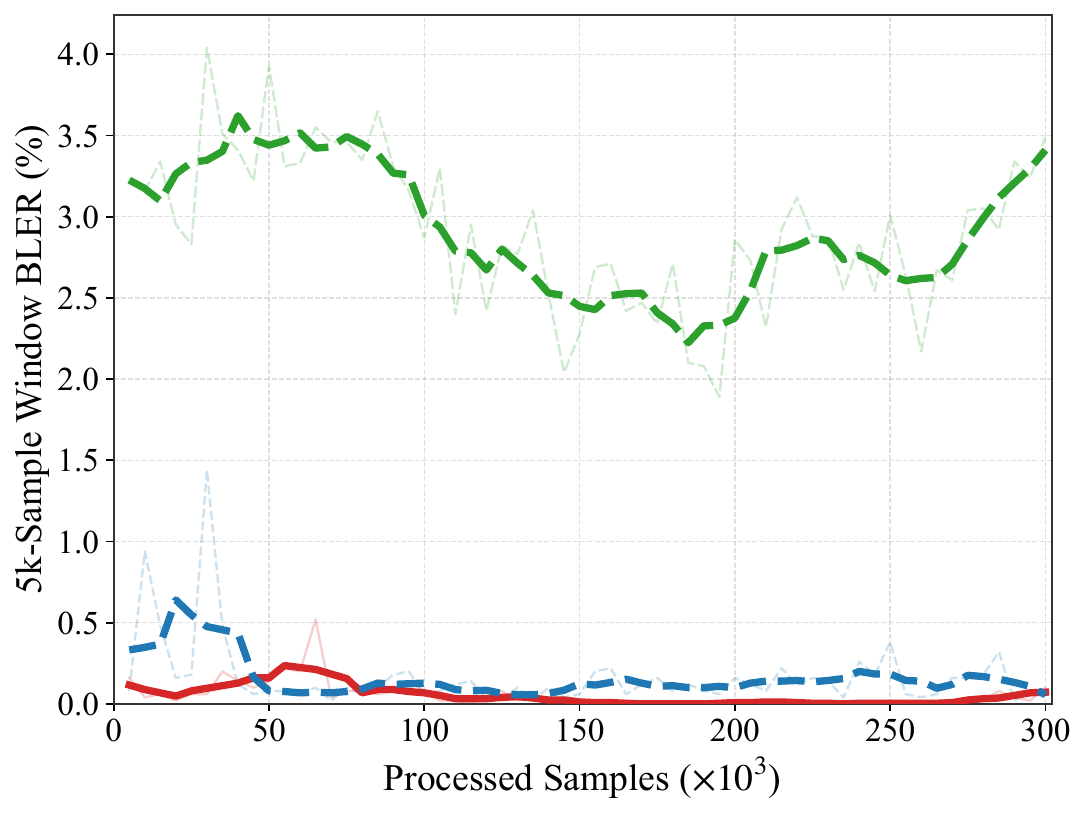}
			\caption{}
			\label{fig:with_interference}
		\end{subfigure}
		\captionsetup{font={small}, justification=raggedright}
		\caption{Over-the-air block error rate over processed samples:
			(a) without interference and (b) with interference. Thin curves are raw
			5k-sample-window measurements; thick curves are five-window moving averages.}
		\label{Validation}
	\end{figure*}
	\subsubsection{Portable Testbed and Measurement Dataset}
	We implement the transmitter and receiver as portable SDR nodes and deploy the
	frozen WGPT backbone configured in Table~\ref{tab:pretraining_details} in the
	receiver chain. The platform operates at \(5\) GHz with a sampling rate of
	\(7.68\) megasamples per second (MS/s). The transmitted waveform employs
	rate-\(1/2\)-coded 16-ary quadrature amplitude modulation (16-QAM) and OFDM with
	a 256-point fast Fourier transform (FFT), \(30\)-kHz subcarrier spacing, 120
	active subcarriers,\footnote{The pretrained WGPT backbone can be directly
	transferred to the 120-subcarrier system configuration because its frequency
	patch size $f_p=6$ exactly divides 120, requiring neither padding nor
	truncation.} and 14 OFDM symbols per slot. Demodulation reference signal
	(DMRS) pilots are inserted at symbols 0 and 11 with a frequency-domain spacing
	of six subcarriers. After synchronization, FFT, and resource-element demapping,
	sparse LS estimates at the DMRS positions are provided to DFT-LMMSE, final-layer
	WGPT, or WGPT+RAFC to reconstruct the full time--frequency CSI. Each estimate is
	evaluated using the same MIMO equalization, demodulation, decoding, and cyclic
	redundancy check (CRC) chain, allowing the resulting BLER differences to be
	attributed to the channel estimator. As shown in Fig.~\ref{Testbed}(a), the
	portable nodes are relocated among office and corridor locations to collect
	channels with different link distances, blockage conditions, and multipath
	characteristics. In Fig.~\ref{Testbed}(b), a separate portable transmitter emits
	a random signal in the same frequency band over half of the system bandwidth,
	thereby introducing structured partial-band co-channel interference.

	\subsubsection{End-to-End BLER Results}
	BLER is useful for determining whether RAFC merely improves an offline
	regression metric or actually increases receiver robustness. Fig.~\ref{Validation} reports the resulting BLER. Without interference, all three
	receivers remain below 0.15\% after smoothing. This small separation is expected:
	in the high-reliability regime, all estimates lie mostly within the decoder's
	tolerance, producing a floor effect even though WGPT+RAFC is generally less
	variable than final-layer WGPT. The benign case therefore verifies that routing
	does not sacrifice nominal link performance, but it is not the most discriminative
	test of the estimators.

	The mechanism becomes visible after the interferer is activated. DFT-LMMSE stays
	between approximately 2.3\% and 3.6\% BLER because its second-order/noise model
	does not explicitly represent colored co-channel contamination; interpolation can
	propagate a corrupted pilot observation across neighboring resource elements.
	Final-layer WGPT is more robust but exhibits larger fluctuations, consistent with
	a globally aggregated representation suppressing interference at the cost of
	occasionally losing channel-specific detail. WGPT+RAFC remains close to zero over
	most of the trace because it can combine the instantaneous evidence retained by
	shallow layers with the interference-tolerant context of intermediate and deep
	layers. The measured result therefore closes the experimental argument: adaptive
	depth composition is not only correlated with lower offline NMSE, but specifically
	protects the decoder when the observation corruption violates the assumptions of
	a fixed model-based estimator.

	\section{Conclusion}\label{Sec6}
	In this paper, we proposed a unified adaptive feature-composition framework for adapting frozen WFMs to heterogeneous downstream tasks. Its central component, RAFC, uses compact descriptors to generate task- and sample-dependent routing weights and combines selected hidden states without modifying the shared backbone. This common adaptation interface remains applicable across different WFM architectures and downstream objectives while introducing only a small number of trainable parameters.
	
	Experiments on channel estimation, channel prediction, beam prediction, and localization demonstrate consistent gains over final-layer-only adaptation, while the learned routing profiles reveal distinct task- and sample-dependent layer preferences. The ablation results further show that spaced layer selection and lightweight feature summarization provide a favorable balance between performance and adaptation cost. Finally, evaluation on the portable SDR platform confirms that the representation gains translate into improved end-to-end receiver robustness across indoor locations and under controlled interference. Overall, RAFC provides a lightweight and reusable interface through which heterogeneous functions in mobile systems can share a frozen WFM backbone.

	\bibliographystyle{IEEEtran} 
	\bibliography{bib1571028021}
	
\end{document}